\documentclass{article}
\usepackage{arxiv}

\usepackage[]{hyperref}
\hypersetup{colorlinks,citecolor=blue}

\usepackage{graphicx}
\usepackage{amsopn}
\usepackage{amssymb}
\usepackage{amsmath}
\usepackage{algorithm}
\usepackage[noend]{algpseudocode}
\usepackage{array}
\usepackage{float}
\usepackage[caption=false,font=footnotesize]{subfig}

\usepackage{colortbl}
\definecolor{myred}{rgb}{0.8,0,0}
\definecolor{myblue}{rgb}{0,0,0.8}

\title{Reinforcement learning with human advice: a survey.}%

\author{
Anis Najar \\
  Laboratoire de Neurosciences Cognitives Computationnelles (LNC2)\\
  INSERM U960, Paris, France\\
  \texttt{anis.najar@ens.fr} \\  
	\And
 Mohamed Chetouani \\
  Institute for Intelligent Systems and Robotics, \\
  Sorbonne Universit\'{e}, CNRS UMR 7222, Paris, France\\
}

\begin{document}
\maketitle

\begin{abstract}
In this paper, we provide an overview of the existing methods for integrating human advice into a Reinforcement Learning process. 
We first propose a taxonomy of the different forms of advice that can be provided to a learning agent.
We then describe the methods that can be used for interpreting advice when its meaning is not determined beforehand.
Finally, we review different approaches for integrating advice into the learning process.

\end{abstract}

%		History, goal
Teaching a machine through natural interaction is an old idea dating back to the foundations of AI, as it was already stated by Alan Turing in 1950: \textit{"It can also be maintained that it is best to provide the machine with the best sense organs that money can buy, and then teach it to understand and speak English. That process could follow the normal teaching of a child. Things would be pointed out and named, etc."} \cite{turing_computing_1950}.
Since then, many efforts have been made for endowing robots and artificial agents with the capacity to learn from humans in a natural and unconstrained manner \cite{chernova_robot_2014}.
However, designing human-like learning robots still raises several challenges regarding their capacity to adapt to different teaching strategies and their ability to take advantage of the variety of teaching signals that can be produced by humans \cite{vollmer_pragmatic_2016}.

%		many types of inputs, but lack of consensus in terminology
The interactive machine learning literature references a plethora of teaching signals such as instructions \cite{pradyot_integrating_2012,najar_interactively_20}, demonstrations \cite{argall_survey_2009} and feedback \cite{knox_interactively_2009,najar_training_2016}.
These signals can be categorized in several ways depending on what, when, and how they are produced.
For example, a common taxonomy is to divide interactive learning methods into three groups: learning from advice, learning from evaluative feedback (or critique) and learning from demonstration \cite{knox_interactively_2009,judah_reinforcement_2010,knox_understanding_2011}.
While this taxonomy is commonly used in the literature, it is not infallible as these categories can overlap. 
For example, in some papers, evaluative feedback is considered as a particular type of advice \cite{judah_reinforcement_2010,griffith_policy_2013}.
In more rare cases, demonstrations \cite{lin1992self,whitehead_complexity_1991} were also referred to as advice \cite{maclin_creating_1996,maclin_giving_2005}.
The definition of advice in the literature is relatively vague with no specific constraints on what type of input can be provided to the learning agent.
For example, it has been defined as \textit{"concept definitions, behavioral constraints, and performance heuristics"} \cite{hayes-roth_advice_1981}, or as \textit{"any external input to the control algorithm that could be used by the agent to take decisions about and modify the progress of its exploration or strengthen its belief in a policy"} \cite{pradyot_beyond_2011}.
Although more specific definitions can be found, such as \textit{"suggesting an action when a certain condition is true"} \cite{knox_interactively_2009}, in other works advice also represents state preferences \cite{utgoff_two_1991}, action preferences \cite{maclin_giving_2005}, constraints on action values \cite{maclin_knowledge-based_2005,torrey_advice_2008}, explanations \cite{Krening2017}, instructions \cite{clouse_teaching_1992,maclin_creating_1996,kuhlmann_guiding_2004,rosenstein_supervised_2004}, feedback \cite{judah_reinforcement_2010,griffith_policy_2013,celemin_coach:_2019}, or demonstrations \cite{maclin_creating_1996,lin1992self,whitehead_complexity_1991}. 
In some papers, the term feedback is used as a shortcut for evaluative feedback \cite{thomaz_reinforcement_2006,leon_teaching_2011,griffith_policy_2013, knox_training_2013,loftin_learning_2016}. 
However, the same term is sometimes used to refer to corrective feedback \cite{argall_teacher_2011}.
While these two types of feedback, evaluative and corrective, are sometimes designated by the same label, they are basically different.
The lack of consensus about the terminology in the literature makes all these concepts difficult to disentangle, and represents an obstacle towards establishing a systematic understanding of how these teaching signals relate to each other from a computational point of view.
%		goal of survey: clarify terminology through taxonomy & review inputs from computational perspective
The goal of this survey is to clarify some of the terminology used in the interactive machine learning literature by providing a taxonomy of the different forms of advice, and to review how these teaching signals can be integrated into a Reinforcement Learning (RL) process \cite{sutton_reinforcement_1998}.
We do not cover learning from demonstration (LfD) in this survey, since demonstrations have some specificity with respect to other types of teaching signals, and comprehensive surveys on this topic already exist \cite{argall_survey_2009,chernova_robot_2014}.

%		focus on RL
Although the methods we cover belong to various mathematical frameworks, we mainly focus on the RL perspective.
We equivalently use the terms of "agent", "robot" and "system", by making abstraction of the support over which the RL algorithm is implemented.
%shaping
Throughout this paper, we use the term "shaping" to refer to the mechanism by which advice is integrated into the learning process. 
Although this concept has been mainly used within the RL literature as a method for accelerating the learning process by providing the learning agent with intermediate rewards \cite{gullapalli_shaping_1992,singh_transfer_1992,dorigo_robot_1994,knox_interactively_2009,judah_imitation_2014,cederborg_policy_2015}, the general meaning of shaping is equivalent to training, which is to make an agent's \textit{"behavior converge to a predefined target behavior"} \cite{dorigo_robot_1994}. 
%In this survey, the term shaping is employed in its general meaning as influencing a learning agent towards a desired behaviour.
%Technically, it qualifies the method used for integrating human teaching signals into an agent's policy.

%So here advice denotes any information that can be communicated by a human teacher to a RL agent in order to modify its behaviour.  

%		Outline
The paper is organized as follows.
We first introduce some background about Reinforcement Learning in Section \ref{RL}.
We then provide an overview of the existing methods for integrating human advice into a RL process in Section \ref{RL_inputs}.
The different methods are discussed in in Section \ref{Discussion}, before concluding in Section \ref{Conclusion}.

%TODO: say it is not exhaustive survey

\section{Reinforcement Learning}
\label{RL}
%		MDP
Reinforcement Learning refers to family of problems where an autonomous agent has to learn a sequential decision-making task \cite{sutton_reinforcement_1998}. 
These problems are generally represented as Markov Decision Process (MDP), defined as a tuple $<S,A,T,R,\gamma>$. 
$S$ represents the state-space over which the problem is defined and $A$ is the set of actions the agent is able to perform on every time-step.
$T: S \times A \to Pr(s'|s,a)$ defines a state-transition probability function, where $Pr(s'|s,a)$ represents the probability that the agent transitions from state $s$ to state $s'$ after executing action $a$.
$R: S \times A \to \mathbb{R}$ is a reward function that defines the reward $r(s,a)$ that the agent gets for performing action $a$ in state $s$.
When at time $t$, the agent performs an action $a_t$ from state $s_t$, it receives a reward $r_{t}$ and transitions to state $s_{t+1}$. 
The discount factor, $\gamma$, represents how much future rewards are taken into account for the current decision.

%		policy, value functions
The behaviour of the agent is represented as a policy $\pi$ that defines the probability to select each action in every state:  $\forall s \in S$, $\pi(s) = \{\pi(s,a);a \in A\} = \{Pr(a|s);a \in A\}$.
The quality of a policy is measured by the amount of rewards it enables the agent to collect over the long run.
The expected amount of cumulative rewards, when starting from a state $s$ and following a policy $\pi$, is given by the state-value function and is written:

\begin{equation} \label{eq:state-value}
V^{\pi}(s) = \sum\limits_a{\pi(s,a)[R(s,a)+ \gamma \sum\limits_{s'}}{Pr(s'|s,a) V^{\pi}(s')}].
\end{equation}

Another form of value function, called action-value function and noted $Q^{\pi}$, provides more directly exploitable information than $V^{\pi}$ for decision-making, as the agent has direct access to the value of each possible decision:

\begin{equation} \label{eq:action-value}
Q^{\pi}(s,a) = R(s,a)+ \gamma \sum\limits_{s'}{Pr(s'|s,a) V^{\pi}(s')} \quad; \forall s \in S, a \in A.
\end{equation}

%		algorithms
To optimize its behaviour, the agent must find the optimal policy $\pi^*$ that maximizes $V^{\pi}$ and $Q^{\pi}$.
%The state and action value functions of $\pi^*$ are noted respectively $V^*$ and $Q^*$.
When both the reward and transition functions are unknown, the optimal policy must be learnt from the rewards the agent obtains by interacting with its environment using a Reinforcement Learning algorithm.
RL algorithms can be decomposed into three categories: value-based, policy-gradient and Actor-Critic \cite{sutton_reinforcement_1998}. 
% such as Q-learning \cite{watkins_q-learning_1992}, SARSA \cite{sutton1996generalization} and Actor-Critic \cite{barto_neuronlike_1983}.
%There exist two main families of RL algorithms: model-free and model-based.
%In model-free RL, the agent learns from the interactions with the environment by trial-and-error, without building a representation of the transition function $T$.
%By contrast, with model-based RL, the agent learns a model of the transition function $T$ and/or the reward function $R$, and uses it for deriving the optimal policy.
%Model-free RL can be further decomposed into three groups: value-based, policy-gradient and Actor-Critic \cite{sutton_reinforcement_1998}. 

\textbf{Value-based RL:}
In value-based RL, the optimal policy is obtained by iteratively optimizing the value-function.
Examples of value-based algorithms include Q-learning \cite{watkins_q-learning_1992} and SARSA \cite{sutton1996generalization}.

%		Q-learning
In Q-learning, the action-value function of the optimal policy $\pi^*$ is computed iteratively.
On every time-step $t$, when the agent transitions from state $s_t$ to state $s_{t+1}$ by performing an action $a_t$, and receives a reward $r_{t}$, the Q-value of the last state-action pair is updated using: 

\begin{equation} \label{eq:q-learning}
Q(s_t,a_t) \gets  Q(s_t,a_t)+ \alpha [r_t + \gamma \max_{a' \in A} Q(s_{t+1},a') - Q(s_t,a_t)],
\end{equation}

where $\alpha \in [0, 1]$ is a learning rate.

At decision time, the policy $\pi$ can be derived from the Q-function using different action-selection strategies. 
The \textit{$\epsilon$-greedy} action-selection strategy consists in selecting most of the time the optimal action with respect to the Q-function, $a_t= \max_{a \in A} Q(s_{t},a)$, and selecting with a small probability $\epsilon$ a random action. 
With the \textit{softmax} action-selection strategy, the policy $\pi$ is derived at decision-time by computing a softmax distribution over the Q-values:

\begin{equation} \label{eq:q-softmax}
\pi(s,a) = Pr(a_t=a|s_t = s) = \frac{e^{Q(s,a)}}{\sum_{b \in A} e^{Q(s,b)}}.
\end{equation}

%	SARSA
The SARSA algorithm is similar to Q-learning, with one difference at the update function of the Q-values: 

\begin{equation} \label{eq:sarsa}
Q(s_t,a_t) \gets  Q(s_t,a_t)+ \alpha [r_t + \gamma Q(s_{t+1},a_{t+1}) - Q(s_t,a_t)],
\end{equation}

where $a_{t+1}$ is the action the agent selects at time-step $t+1$. 
At decision time, the same action-selection strategies can be implemented as for Q-learning.

\textbf{Policy-gradient RL:}
In contrast to value-based RL, policy-gradient methods do not compute a value function \cite{williams1992simple}.
Instead, the policy is directly optimized from the perceived rewards.
In this approach, the policy $\pi$ is controlled with a set of parameters $w \in \mathbb{R}^{n}$, such that $\pi_{w}(s,a)$ is differentiable in $w; \forall s \in S , a\in A$.
For example, $w$ can be defined so that $w(s,a)$ reflects the preference for taking a action in a given state, by expressing the policy as a softmax distribution over the parameters:

\begin{equation} \label{eq:actor}
\pi_w(s,a) = Pr(a_t=a|s_t = s) = \frac{e^{w(s,a)}}{\sum_{b \in A} e^{w(s,b)}}.
\end{equation}

A learning iteration is composed of two stages.
First, the agent estimates the expected returns, $G$, by sampling a set of trajectories.
Then, the policy $\pi_{w}$ is updated using the gradient of the expected returns with respect $w$.
For example, in the REINFORCE algorithm \cite{williams1992simple}, a trajectory of $T$ time-steps is first sampled from one single episode. 
Then, for every time-step $t$ of the trajectory, the return $G$ is computed as $G \gets \sum_{k=t+1}^{T} \gamma^{k-t-1} r_t$, and the policy parameters are updated with:

\begin{equation} \label{eq:pgrl}
w \gets w + \gamma^t G \nabla_w \ln \pi_w(a_t|s_t).
\end{equation}

%		Actor-critic
\textbf{Actor-Critic RL:} 
Actor-Critic architectures constitute a hybrid approach between value-based and policy-gradient methods, by computing both the policy (the actor) a value function (the critic).
The actor can be represented as a paramterized softmax distribution as in Eq. \ref{eq:actor}.
The critic computes a value function that is used for evaluating the the actor. 
%Typically, it is a state-value function, but an action-value function can also be used. 
The reward $r_{t}$ received at time $t$ is used for computing a temporal difference (TD) error:

\begin{equation} \label{eq:td-error}
\delta_t = r_{t} + \gamma V(s_{t+1}) - V(s_t).
\end{equation}

The TD error is then used for updating both the critic and the actor, using respectively Equations (\ref{eq:ac-vu}) and (\ref{eq:ac-pu}):

\begin{equation} \label{eq:ac-vu}
V(s_t) \leftarrow V(s_t) + \alpha \delta_t,
\end{equation}

\begin{equation} \label{eq:ac-pu}
w(s_t,a_t) \leftarrow w(s_t,a_t) + \beta \delta_t,
\end{equation}

where $\alpha \in [0, 1]$ and $\beta \in [0, 1]$ are two learning rates.
A positive TD error increases the probability of selecting $a_t$ in $s_t$, while a negative TD error decreases it. 

% %		Model-based
% \textbf{Model-based RL:} 
% One example of model-based RL is given by the Dyna-Q algorithm \cite{sutton1990integratedarchitecturesforlearning}.

% TODO: describe the algorithm and introduce planning

\section{Reinforcement Learning with human advice}
\label{RL_inputs}
%\subsection{Intro}
%5-Steps advice-taking
In one of the first papers of Artificial Intelligence, John McCarthy described an \textit{"Advice Taker"} system that could learn by being told \cite{mccarthy_programs_1959}.
This idea was then elaborated in \cite{hayes-roth_knowledge_1980,hayes-roth_advice_1981}, where a general framework for learning from advice was proposed.
This framework can be summarized in the following five steps \cite{cohen_handbook_1982,maclin_creating_1996}: \\
1. Requesting or receiving the advice.\\ 
2. Converting the advice into an internal representation.\\ 
3. Converting the advice into a usable form (Operationalization).\\ 
4. Integrating the reformulated advice into the agent's knowledge base.\\ 
5. Judging the value of the advice.\\ 

%applies to any type of input, also feedback and demonstrations
%Although this framework has been designed for advice-taking systems, it can be generalized to any type of human input.
%description
%step1. providing human inputs
The first step describes how human advice can be provided to the system.
Different forms of advice can be distinguished based on this criterion. 
%Representation & interpretation
Step 2 refers to the encoding the perceived advice into an internal representation.
Most of existing advice-taking systems assume that the internal representation of advice is predetermined by the system designer. 
However, some recent works tackle the problem of letting the system learn how to interpret raw advice, in order to make the interaction protocol less constraining for the human teacher \cite{vollmer_pragmatic_2016}.
%As interpreting advice is a relatively recent research question, and most of existing methods predefine the meaning of teaching signals, we cover this aspect at the end of each section.
%shaping
Steps 3 to 5 describe how human advice can be used by the agent for learning. 
These three steps are often confounded into one single process, that we call shaping, which consists in integrating advice into the agent's learning process.

In the remainder of this section, we first propose a taxonomy of different categories of advice based on how they can be provided to the system (step 1).
Then we detail how advice can be interpreted (step 2).
Finally, we present how advice can be integrated into a RL process (steps 3 to 5).

%We first propose a taxonomy of the different forms of advice that can be provided to a learning agent.
%We then describe the methods that can be used for interpreting advice when its meaning is not determined beforehand.
%Finally, we review different approaches for integrating advice into the learning process.

%but there are differences
%However, we have different kinds of inputs with some specificities, especially in terms of what type of information is provided to the agent, and how this information will be integrated to the system

%so we introduce a taxonomy
%Here we provide a taxonomy of human inputs or teaching signals that can be used for guiding a RL agent.

%These three sections follow the same structure based on three main aspects:
%how advice can be provided to the agent, how it can be integrated into the learning process, and how it can be interpreted by the agent if their meaning is not determined beforehand.
\subsection{Providing advice}
\label{taxonomy}
The means by which teaching signals can be communicated to a learning agent vary.
They can be provided via natural language \cite{kuhlmann_guiding_2004,cruz_interactive_2015,PaleologueMPC18}, computer vision \cite{atkeson_learning_1997,najar_interactively_20}, hand-written programs \cite{maclin_creating_1996,maclin_giving_2005,maclin_knowledge-based_2005,torrey_advice_2008}, artificial interfaces \cite{abbeel_autonomous_2010,suay_effect_2011, knox_training_2013}, or physical interaction \cite{lozano-perez_robot_1983,akgun_trajectories_2012}.
% %advice vs demonstrations
Despite the variety of communication channels, we can distinguish two main categories of teaching signals based on how they are produced: advice and demonstration.
Even though advice and demonstration can share the same communication channels, like computer vision \cite{atkeson_learning_1997,najar_interactively_20} and artificial interfaces \cite{abbeel_autonomous_2010,suay_effect_2011, knox_training_2013}, they are fundamentally different from each other in that demonstration requires the task to be executed by the teacher (demonstrated), while advice does not.
In rare cases, demonstration \cite{lin1992self,whitehead_complexity_1991} has been referred to as advice \cite{maclin_creating_1996,maclin_giving_2005}.
However, it is more common to consider demonstration and advice as two distinct and complementary approaches for interactive learning \cite{dillmann2000learning,argall2008learning,knox_interactively_2009,judah_reinforcement_2010,knox_understanding_2011}.
Based on this distinction, we define advice as \textit{teaching signals that can be communicated by the teacher to the learning system without executing the task}. 

\begin{figure}[h!]
\begin{center}
\includegraphics[trim=3.5cm 4cm 1.5cm 1.5cm, clip=true,scale=0.8]{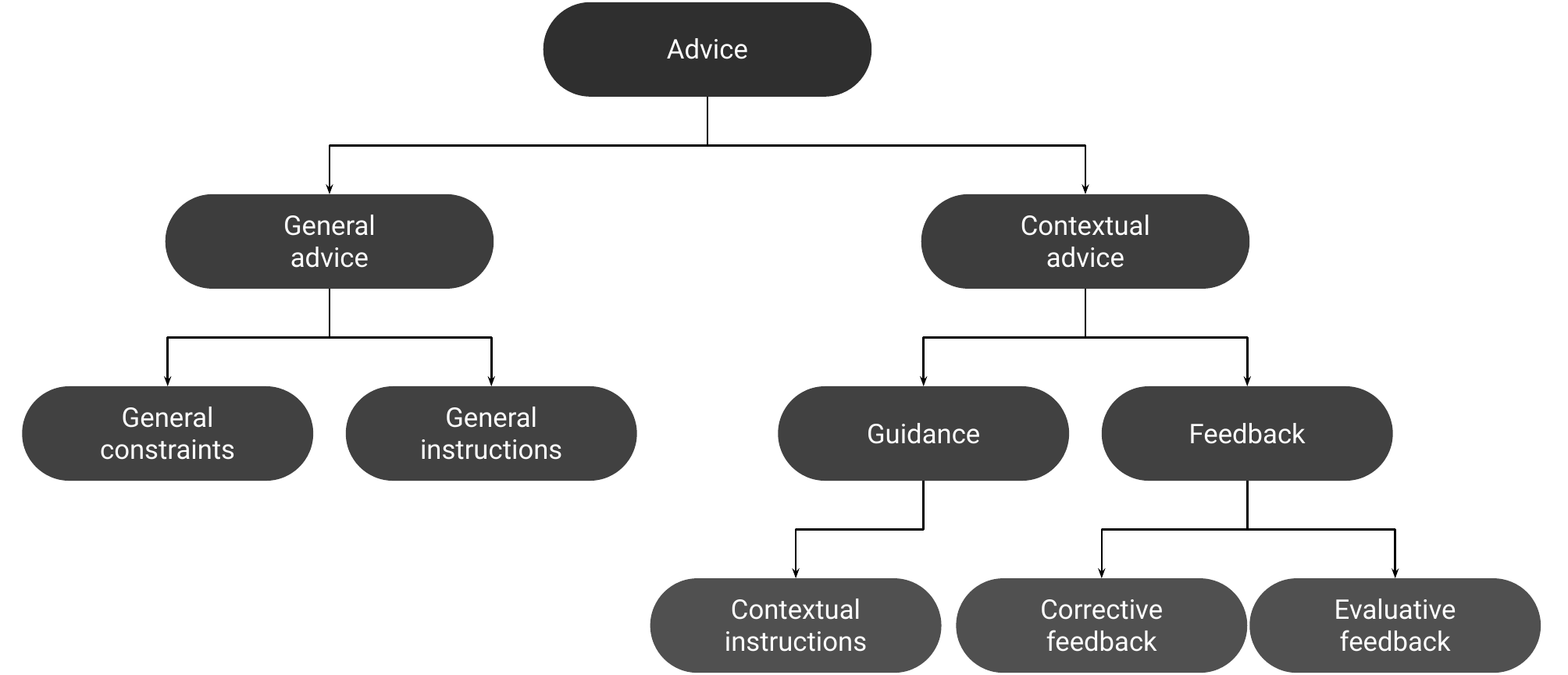}
\caption{Taxonomy of advice.} 
\label{fig:structure}
\end{center}
\end{figure}

%advice
We mainly distinguish two forms of advice depending on how it is provided to the system: \textit{general advice} and \textit{contextual advice} (Fig. \ref{fig:structure}, Table \ref{tab:structure}). %While demonstrations are generally provided offline (\textit{i.e.,} prior to the learning process) in a batch fashion, where several task executions are first collected from the teacher then communicated to the learning system, we distinguish two types of advice: \textit{general} and \textit{contextual}.
%general advice
%Advice can be provided in two different forms: \textit{general} and \textit{contextual} .
\textit{General advice} can be communicated to the system, non-interactively, prior to the learning process (offline).
This type of advice represents information about the task that do not depend on the context in which they are provided. 
They are self-sufficient in that they include all the required information for being converted into a usable form (operationalization). 
%Examples include detailed action plans, task specifications and behavioural constraints.
Examples include specifying general constraints about the task and providing general instructions about the desired behaviour.
%contextual advice
\textit{Contextual advice}, on the other hand, is context-dependent, in that the communicated information depends on the current state of the task.
So, unlike \textit{general advice}, it must be provided interactively along the task \cite{knox_interactively_2009,celemin_coach:_2019,najar_interactively_20}.
%batch/interactive
\textit{Contextual advice} can also be provided in an offline fashion, with the teacher interacting with previously recorded task executions by the learning agent \cite{judah_reinforcement_2010,argall_teacher_2011}. 
Even in this case, each piece of advice has to be provided at a specific moment of the task execution.
Examples of \textit{contextual advice} include evaluative feedback  \cite{knox_interactively_2009,najar_training_2016}, corrective feedback \cite{argall_teacher_2011,celemin_coach:_2019}, guidance \cite{thomaz_reinforcement_2006,suay_effect_2011} and contextual instructions \cite{clouse_teaching_1992,rosenstein_supervised_2004,pradyot_instructing_2012,najar_interactively_20}.

\textbf{General advice}
%\label{general-advice}
%	General constraints
Advice can be used by the human teacher to provide the agent with general information about the task prior to the learning process. 
These information can be provided to the system in a written form \cite{hayes-roth_knowledge_1980,maclin_creating_1996,kuhlmann_guiding_2004,branavan_reinforcement_2009,vogel_learning_2010}.

General advice can specify \textit{general constraints} about the task such as domain concepts, behavioural constraints and performance heuristics.
For example, the first ever implemented advice-taking system relied on general constraints that were written as LISP expressions, to specify concepts, rules and heuristics for a card-playing agent \cite{hayes-roth_advice_1981}.
%When the executed advice lead to unexpected or unfavorable consequences, learning was triggered by correcting the advice and refining the knowledge base through predefined learning rules.
%	State preferences
%TODO:	goal specification
%TODO:	Action preferences
%	Providing general instructions

A second form of general advice, \textit{general instructions}, explicitly specifies to the agent what actions to perform in different situations.
It can be provided either in the form of \textit{if-then} rules \cite{maclin_creating_1996,kuhlmann_guiding_2004}, or as detailed action plans describing the step-by-step sequence of actions that should be performed in order to solve the task \cite{branavan_reinforcement_2009,vogel_learning_2010}.
Action plans can be seen as a sequence of low-level or high-level \textit{contextual instructions} (cf. definition below).
For example, a sequence like (e.g. \textit{"Click start, point to search, and then click for files or folders."}), can be decomposed into a sequence of three low-level \textit{contextual instructions} \cite{branavan_reinforcement_2009}.

%However, unlike \textit{if-then} rules, state information in general instructions can be implicit. 
%For example, they can be implicitly formulated within the expression of the action (e.g. \textit{"Click start, point to search, and then click for files or folders."}) \cite{branavan_reinforcement_2009}.

\textbf{Contextual advice:}
%\label{contextual-advice}
%\paragraph{Contextual instructions}
%	Providing contextual instructions
In contrast to \textit{general advice}, a \textit{contextual advice} depends on the state in which it is provided. 
To use the terms of the advice-taking process, a part of the information that is required for operationalization is implicit, and must be inferred by the learner from the current context.
%More specifically, the condition part of each instruction is not explicitly communicated by the teacher, but must be inferred by the learner from the current context.
Consequently, \textit{contextual advice} must be progressively provided to the learning agent along the task. 
Contextual advice can be divided into two main categories: guidance and feedback. 
Guidance informs about future actions whereas feedback informs about past ones. 

%\paragraph{Guidance}
\textbf{Guidance:} Guidance is a term that is encountered in many papers and has been made popular by the work of Thomaz \cite{thomaz_socially_2006} about Socially Guided Machine Learning . 
In the broad sense, guidance represents the general idea of guiding the learning process of an agent.
In this sense, all interactive learning methods can be considered as a form of guidance.
A bit more specific definition of guidance is when human inputs are provided in order to bias the exploration strategy \cite{thomaz_learning_2009}.
For instance, in \cite{subramanian_exploration_2016}, demonstrations were provided in order to teach the agent how to explore interesting regions of the state space. 
In \cite{chu_learning_2016}, kinesthetic teaching was used for guiding the exploration process for learning object affordances.
In the most specific sense, guidance constitutes a form of advice that consists in suggesting a limited set of actions from all the possible ones \cite{suay_effect_2011,thomaz_reinforcement_2006}.

\textbf{Contextual instructions:} One particular type of guidance is to suggest only one action to perform. 
We refer to this type of advice as \textit{contextual instructions}.
For example, in \cite{cruz_interactive_2015}, the authors used both terms of advice and guidance for referring to contextual instructions.
Contextual instructions can be either low-level or high-level \cite{branavan_reading_2010}.
Low-level instructions indicate the next action to perform \cite{grizou_robot_2013}, whereas high-level instructions indicate a more extended goal without explicitly specifying the sequence of actions that should be executed \cite{macglashan_translating_2014}. 
High-level instructions were also referred to as commands \cite{macglashan_translating_2014,tellex_learning_2014}. %est-ce pertinent ?
In RL terminology, high-level instructions would correspond to performing \textit{options} \cite{sutton1999between}.
Contextual instructions can be provided for example through speech \cite{grizou_robot_2013}, gestures \cite{najar_interactively_20} or myoelectric (EMG) interfaces \cite{mathewson_simultaneous_2016}.

\textbf{Feedback:} We distinguish two main forms of feedback: evaluative and corrective.
Evaluative feedback, also called critique, consists in evaluating the quality of the agent's actions \cite{knox_interactively_2009,judah_reinforcement_2010} .
Corrective feedback, also called instructive feedback, implicitly implies that the performed action is wrong \cite{argall_teacher_2011,celemin_coach:_2019}. 
However, it goes beyond simply criticizing the performed action, by informing the agent about the correct one. %which action it should have performed. 
%Both forms of feedback can be provided either interactively after each performed action, or a posteriori to the task execution, in a batch fashion.
%While corrective feedback is used in several works, much more emphasis has been put on evaluative feedback, especially as a standalone training method. 

\textbf{Corrective feedback:}
%\paragraph{Corrective feedback}
%We distinguish two main forms of feedback: evaluative and corrective.
%Evaluative feedback, also called critique, consists in evaluating the quality of the agent's actions.
Corrective feedback can be either a corrective instruction \cite{chernova_interactive_2009} or a corrective demonstration \cite{nicolescu_natural_2003}. 
%These two forms of corrective feedback are also called respectively corrective instructions and corrective demonstrations.
%In the latter case, we also talk about corrective demonstrations.
The main difference with instructions (resp. demonstrations) is that they are provided after an action (resp. a sequence of actions) is executed by the agent, not before.
So, operationalization is made with respect to the previous state instead of the current one.

So far, corrective feedback has been mainly used for augmenting LfD systems \cite{nicolescu_natural_2003,chernova_interactive_2009,argall_teacher_2011}.
For example, in \cite{chernova_interactive_2009}, while the robot is reproducing the provided demonstrations, the teacher could interactively rectify any incorrect action.
In \cite{nicolescu_natural_2003}, corrective demonstrations were delimited by two predefined verbal commands that were pronounced by the teacher. 
In \cite{argall_teacher_2011}, the authors presented a framework based on \textit{advice-operators}, allowing a teacher to correct entire segments of demonstrations through a visual interface.
Advice-operators were defined as numerical operations that can be performed on state-action pairs. 
The teacher could choose an operator from a predefined set, and apply it to the segment to be corrected.
In \cite{celemin_coach:_2019}, the authors took inspiration from advice-operators to propose learning from corrective feedback as a standalone method, contrasting with other methods for learning from evaluative feedback such as TAMER \cite{knox_interactively_2009}.

\textbf{Evaluative feedback:}
%Providing evaluative feedback
Teaching an agent by evaluating its actions is an alternative solution to the standard RL approach.
Evaluative feedback can be provided in different forms: a scalar value $f \in [-1,1]$ \cite{knox_interactively_2009}, a binary value $f \in \{-1,1\}$ \cite{thomaz_reinforcement_2006-1,najar_interactively_20}, a positive reinforcer $f \in \{"Good!","Bravo!"\}$ \cite{kaplan_robotic_2002}, or a categorical information $f \in \{Correct, Wrong\}$ \cite{loftin_learning_2016}. 
These values can be provided through buttons \cite{kaplan_robotic_2002,suay_effect_2011, knox_training_2013}, speech \cite{kim_learning_2007,grizou_robot_2013}, gestures \cite{najar_interactively_20}, or electroencephalogram (EEG) signals \cite{grizou_calibration-free_2014}.

Another form of evaluative feedback is to provide preferences between demonstrated trajectories \cite{sadigh2017active,christiano2017deep,cui2018active}.
Instead of critiquing one single action or a sequence of actions, the teacher provides a ranking for demonstrated trajectories.
The provided human preferences are then aggregated in order to infer the reward function.
This form of evaluative feedback has been mainly investigated within the LfD community as an alternative to the standard Inverse Reinforcement Learning approach (IRL), by relaxing the constraint for the teacher to provide demonstrations.

\begin{table}[!ht]
\centering
\begin{tabular}[t]{l>{\raggedright\arraybackslash}p{0.7\textwidth}}
\hline
Category	            & References\\
\hline
General constraints	    &  \cite{hayes-roth_advice_1981,mangasarian_knowledge-based_2004,kuhlmann_guiding_2004,maclin_giving_2005,maclin_knowledge-based_2005,torrey_advice_2008}\\
General instructions    & \cite{maclin_creating_1996,kuhlmann_guiding_2004,branavan_reinforcement_2009,vogel_learning_2010,branavan_reading_2010}\\
Guidance                & \cite{thomaz_socially_2006,thomaz_learning_2009,suay_effect_2011,subramanian_exploration_2016,chu_learning_2016}\\
Contextual instructions & \cite{utgoff_two_1991,clouse_teaching_1992,nicolescu_natural_2003,rosenstein_supervised_2004,thomaz_robot_2007,rybski_interactive_2007,tenorio-gonzalez_dynamic_2010,branavan_reading_2010,pradyot_integrating_2012,grizou_robot_2013,macglashan_translating_2014,cruz_interactive_2015,mathewson_simultaneous_2016,najar_interactively_20}\\
Corrective feedback     & \cite{nicolescu_natural_2003,chernova_interactive_2009,argall_teacher_2011,celemin_coach:_2019}\\
Evaluative feedback     & \cite{dorigo_robot_1994,colombetti_behavior_1996,isbell_social_2001,kaplan_robotic_2002,thomaz_reinforcement_2006-1,kim_learning_2007,knox_interactively_2009,judah_reinforcement_2010,tenorio-gonzalez_dynamic_2010,lopes_simultaneous_2011,knox_combining_2010,knox_augmenting_2011,knox_reinforcement_2012-1,knox_reinforcement_2012,grizou_robot_2013} \\
                        & \cite{griffith_policy_2013,grizou_interactive_2014,loftin_strategy-aware_2014,ho_teaching_2015,loftin_learning_2016,mathewson_simultaneous_2016,macglashan2017interactive,najar_training_2016,najar_interactively_20} \\
\hline
\end{tabular}
\caption{Types of advice.} 
\label{tab:structure} 
\end{table}

\subsection{Interpreting advice}
\label{interpretation}
%%introduce unalbeled teaching signals!!!

The second step of the advice-taking process stipulates that advice needs to be converted into an internal representation.
%So far, providing \textit{general advice} to a learning system required to write them down using specific programming languages \cite{maclin_creating_1996,mangasarian_knowledge-based_2004,kuhlmann_guiding_2004,maclin_giving_2005,maclin_knowledge-based_2005,torrey_advice_2008}. 
Predefining the meaning of advice by hand-coding the mapping between raw signals and their internal representation, has been widely used in the literature \cite{clouse_teaching_1992,nicolescu_natural_2003,rosenstein_supervised_2004,thomaz_robot_2007,rybski_interactive_2007,chernova_interactive_2009,tenorio-gonzalez_dynamic_2010,lockerd_tutelage_2004,pradyot_instructing_2012,cruz_interactive_2015,celemin_coach:_2019}. %include also feedback and guidance
However, this solution has many limitations.
First, programming the meaning of raw advice signals for new tasks requires expert programming skills, which is not accessible to all human users. 
Second, it limits the possibility for different teachers to use their own preferred signals.

One way to address these limitations is to teach the system how to interpret the teacher's raw advice signals. 
This way, the system would be able to understand advice, that can be expressed for example through natural language or non-verbal cues, without predetermining the meaning of each signal.
In this case, we talk about learning with unlabeled teaching signals \cite{grizou_interactive_2014,najar_interactively_20}.
%The ultimate goal of advice-taking methods is to be able to take advantage of human advice expressed in natural language.%However, natural language understanding still raises many challenges.
To achieve this goal, different approaches have been taken in the literature. 
Table \ref{tab:interpretation} summarizes the literature addressing the question of interpreting advice.
We categorize them according to the type of advice, the communication channel, the interpretation method, and the inputs given to the system for interpretation.

\begin{table}[htb!]
\centering
%\begin{adjustbox}{angle=-90}
\begin{tabular}{lllll}
\hline
Reference 							& Instructions & Channel & Method  & Inputs \\
\hline

\cite{kate_using_2006}  	        & GI            & Text      & SVM           & Demonstration*            \\
\cite{kim_learning_2007}  	        & EFB           & Speech    & kNN           & Binary EFB classes        \\
\cite{chen_learning_2011}  	        & GLI           & Text      & SVM           & Demonstration             \\
\cite{tellex_understanding_2011}  	& GHI           & Text      & Graphical model & Demonstration           \\
\cite{artzi_weakly_2013}  			& GHI           & Text      & Perceptron    & Rewards/demonstration + LM \\
\cite{duvallet_imitation_2013}  	& GLI           & Text      & MCC           & Demonstration + LM \\
\cite{tellex_learning_2014} 		& GHI           & Text      & Gradient descent& Demonstration           \\
\cite{pradyot_integrating_2012}		& CLI           & Gestures  & MLN           & Demonstration*            \\
\cite{lopes_simultaneous_2011}  	& EFB and CFB   & SDI       & IRL           & EFB and CFB               \\
\cite{grizou_robot_2013} 			& EFB or CLI    & Speech    & EM            & Task models               \\
\cite{grizou_interactive_2014}      & EFB           & EEG       & EM            & Task models               \\
\cite{macglashan_translating_2014}  & GHI           & Text      & EM            & Task and language models   \\
\cite{macglashan_training_2014}     & GHI           & Text      & EM            & EFB + LM      \\
\cite{loftin_learning_2016}  	    & EFB           & Buttons   & EM            & Task models               \\
\cite{branavan_reinforcement_2009}	& GLI           & Text      & PGRL          & Rewards                   \\
\cite{branavan_reading_2010}	    & GHI           & Text      & MB-PGRL       & Rewards                   \\
\cite{vogel_learning_2010}			& GLI           & Text      & SARSA         & Demonstration             \\
\cite{najar_socially_2015}          & CLI           & SDI       & XCS           & Rewards                   \\
\cite{najar_social-task_2015}       & CLI           & Gestures  & XCS           & EFB                       \\
\cite{najar_training_2016}          & CLI           & Gestures  & Q-learning    & EFB                       \\
\cite{mathewson_simultaneous_2016}	& CLI           & EMG       & ACRL          & Rewards and/or EFB        \\
\cite{najar_interactively_20}       & CLI           & Gestures  & ACRL          & Rewards and/or EFB        \\

\hline
\end{tabular}
%\end{adjustbox}
\caption{Interpreting advice.GI: General instruction. GLI: general low-level instruction. GHI: general high-level instruction. CLI: contextual low-level instruction. EFB: evaluative feedback. CFB: corrective feedback. SVM: Support Vector Machines. kNN: k-nearest neighbors. MCC: multi-class classification. MLN: Markov Logic Networks. IRL: Inverse Reinforcement Learning. PGRL: policy-gradient RL. MB-PGRL: model-based policy-gradient RL. ACRL: Actor-Critic RL. LM: Language model. *The term demonstration here is taken in the general sense as a trajectory, not necessarily the optimal one.} 
\label{tab:interpretation} 
\end{table}

\textbf{Supervised interpretation:} Some methods relied on interpreters trained with supervised learning methods \cite{kate_using_2006,zettlemoyer_learning_2009,matuszek_learning_2013}. 
For example, in \cite{kuhlmann_guiding_2004}, the system was able to convert general instructions expressed in a constrained natural language into a formal representation using \textit{if-then} rules, by using a parser that was previously trained with annotated data.
In \cite{pradyot_integrating_2012}, two different models of contextual instructions were learned in the first place using Markov Logic Networks (MLN) \cite{domingos2016unifying}, and then used for guiding a learning agent in a later phase.
The most likely interpretation was taken from the instruction model with the highest confidence.
In \cite{kim_learning_2007}, a binary classification of prosodic features was performed offline, before using it to convert evaluative feedback into a numerical reward signal for task learning.

\textbf{Grounded interpretation:} More recent approaches take inspiration from the \textit{grounded language acquisition} literature \cite{mooney_learning_2008}, to learn a model that grounds the meaning of advice into concepts from the task. 
For example, general instructions expressed in natural language can be paired with demonstrations of the corresponding tasks to learn the mapping between low-level contextual instructions and their intended actions \cite{chen_learning_2011,tellex_understanding_2011,duvallet_imitation_2013}.
%Other works rely on demonstrations \cite{}.
In \cite{macglashan_translating_2014}, the authors proposed a model for grounding general high-level instructions into reward functions from user demonstrations.
The agent had access to a set of hypotheses about possible tasks, in addition to command-to-demonstration pairings. 
Generative models of tasks, language, and behaviours were then inferred using Expectation Maximization (EM) \cite{dempster_maximum_1977}.
In addition to having a set of hypotheses about possible reward functions, the agent was also endowed with planning abilities that allowed it to infer a policy according to the most likely task.
The authors extended their model in \cite{macglashan_training_2014}, to ground command meanings in reward functions, using evaluative feedback instead of demonstrations. 
%\cite{peng_language_2015}?

%
In a similar work \cite{grizou_robot_2013}, a robot learned to interpret both low-level contextual instructions and evaluative feedback, while inferring the task using an EM algorithm. 
Contextual advice was interactively provided through speech.
As in \cite{macglashan_training_2014}, the robot knew the set of possible tasks, and was endowed with a planning algorithm allowing it to derive a policy for each possible task.
This model was also used for interpreting evaluative feedback provided through EEG signals \cite{grizou_interactive_2014}. 
In \cite{lopes_simultaneous_2011}, a predefined set of known feedback signals, both evaluative and corrective, were used for interpreting additional signals with IRL \cite{ng_algorithms_2000}.
%Finally, some papers addressed the question of interpreting the teacher's silence, which was referred to as implicit feedback \cite{loftin_learning_2016}. 
%TODO explain how		

\textbf{RL-based interpretation:} A different approach relies on Reinforcement Learning for interpreting advice \cite{branavan_reinforcement_2009,branavan_reading_2010,vogel_learning_2010,mathewson_simultaneous_2016,najar_interactively_20}.
In \cite{branavan_reinforcement_2009}, the authors used a policy-gradient RL algorithm with a predefined reward function to interpret general low-level instructions for a software application.
%, to map textual low-level instructions into actions in a GUI application. 
%Contextual low-level instructions were provided \textit{a priori} as a general instruction, detailing the step-by-step sequence of actions that must be performed. 
This model was extended in \cite{branavan_reading_2010}, to allow for the interpretation of high-level instructions, by learning a model of the environment. %as in model-based RL. 
In \cite{vogel_learning_2010}, a similar approach was used for interpreting general low-level instructions, in a path-following task, using the SARSA algorithm.
The rewards were computed according to the deviation from a provided demonstration.

In \cite{mathewson_simultaneous_2016}, contextual low-level instructions were provided to a prosthetic robotic arm in the form of myoelectric control signals, and interpreted using evaluative feedback with an Actor-Critic architecture. 
In \cite{najar_socially_2015}, a model of contextual low-level instructions was built using the XCS algorithm \cite{butz_algorithmic_2001} in order to predict task rewards, and used simultaneously for speeding-up the learning process.
This model was extended in \cite{najar_social-task_2015}, to predict action values instead of task rewards.
In \cite{najar_training_2016}, interpretation was based on evaluative feedback using the Q-learning algorithm.
In \cite{najar2017shaping}, several methods for interpreting contextual low-level instructions were compared.
Each contextual low-level instruction was defined as a \textit{signal policy} representing a probability distribution over the action-space, in the same way as a RL policy:

\begin{equation}
\pi(i) = \{\pi(i,a); a \in A\} = \{Pr(a|i); a \in A\},
\end{equation}

where $i$ is an observed instruction signal. Two types of interpretation methods were proposed: batch and incremental.
The main idea of batch interpretation methods is to derive a state policy for an instruction signal by combining the policies of every task state in which it has been observed.
Different combination methods were investigated. 
The Bayes optimal solution derives the signal policy by marginalizing the state policies over all the states where the signal has been observed:

\begin{align}
\pi(i,a) = Pr(a|i) &= \sum_{s \in S} Pr(a|s) \times Pr(s|i)\\
         &= \sum_{s \in S} \pi(s,a) \times Pr(i|s) \times Pr(s)/Pr(i),
\end{align}

where $Pr(i|s), Pr(s)$ and $Pr(i)$ represent respectively the probability of observing the signal $i$ in state $s$, the probability of being in state $s$ and the probability of observing the signal $i$.

Other batch interpretation methods were inspired from ensemble methods \cite{wiering_ensemble_2008}, which have been classically used for combining the policies of different learning algorithms.
These methods compute preferences $p(i,a)$ for each action, which are then transformed into a policy using the softmax distribution as in Eq. \ref{eq:actor}.
The Boltzmann Multiplication consists in multiplying the policies:

\begin{equation}
p(i,a)=\prod_{s \in S;i^*(s)=i} \pi(s,a),
\end{equation}

where $i^*(s)$ represents the instruction signal associated to the state $s$. 

The Boltzamm Addition consists in adding the policies:

\begin{equation}
p_t(i,a)=\sum_{s \in S;i^*(s)=i} \pi_t(s,a).
\end{equation}

In Majority Voting, the most preferred interpretation for a signal $i$ is the action that is optimal the most often over all its contingent states:

\begin{equation}
    p(i,a)=\sum_{s \in S;i^*(s)=i} I(\pi^*(s),a),
\end{equation}

where $I(x,y)$ is the indicator function that outputs $1$ when $x=y$ and $0$ otherwise. 

In Rank Voting, the most preferred action for $i$ is the one that has the highest cumulative ranking over all its contingent states:
\begin{equation}
    p(i,a)=\sum_{s \in S;i^*(s)=i} R(s,a),
\end{equation}
where $R(s,a)$ is the rank of action $a$ in state $s$, such that if $a_j$ and $a_k$ denote two different actions and $\pi(s,a_j) \geq \pi(s,a_k)$ then $R(s,a_j) \geq R(s,a_k)$.

Local interpretation methods, on the other hand, incrementally update the meaning of each instruction signal using information from the task learning process such as the rewards, the TD error, or the policy gradient.
With Reward-based Updating, instruction signals constitute the state space for an alternative MDP, that is solved using a standard RL algorithm. 
This approach is similar to the one used in \cite{branavan_reading_2010,branavan_reinforcement_2009,vogel_learning_2010}.
In Value-based Updating, the meaning of instruction is updated with the same amount as the Q-values of its corresponding state:

\begin{equation}
    \delta p_t(i,a_t)=\delta Q(s_t,a_t),
\end{equation}

whereas in Policy-based Updating, it is updated using the policy update:

\begin{equation}
    \delta \pi(i,a_t)=\delta \pi(s_t,a_t).
\end{equation}
These methods were compared using both a reward function and evaluative feedback.
Policy-based Updating presented the best compromise in terms of performance and computation cost.

\subsection{Shaping with advice}
\label{shaping}
%Intro
%	Several ways for a RL agent to learn from advice.
We can distinguish several strategies for integrating advice into a RL system, depending to which stage of the learning process is influenced by the the advice.
The overall RL process can be summarized as follows.
First, the main source of information to a RL agent is the reward function.
In value-based RL, the reward function is used for computing a value function, which is then used for deriving a policy.
In policy-based RL, the policy is directly derived from the reward function without computing any value function.
Finally, the policy is used for decision-making.
Advice can be integrated into the learning process at any of these four different stages: the reward function, the value function, the policy or the decision.

We qualify the methods used for integrating advice as shaping methods.
In the literature, this term has been used exclusively used for evaluative feedback, especially as a technique for providing extra-rewards.
For example, we find different terminologies such as reward shaping \cite{tenorio-gonzalez_dynamic_2010}, interactive shaping \cite{knox_interactively_2009} and policy shaping \cite{griffith_policy_2013,cederborg_policy_2015}. 
In some works, the term shaping is not even adopted \cite{loftin_learning_2016}.
In this survey we generalize this term to all types of advice, by considering the term shaping in its general meaning as influencing a RL agent towards a desired behaviour.
In this sense, all methods for integrating advice into a RL process are considered as shaping methods, especially that similar shaping patterns can be found across different categories of advice.
%In what follows, we propose a categorization of different shaping methods with advice.

We distinguish four main strategies for integrating advice into a RL system: reward shaping, value shaping, policy shaping and decision-biasing, depending on the stage in which advice is integrated into the learning process (cf. Table \ref{tab:shaping}).
Orthogonal to this categorization, we distinguish model-free from model-based shaping strategies.
In model-free shaping, the perceived advice is directly integrated into the learning process, whereas model-based shaping methods build a model of the teacher that is kept in parallel with the agent's own model of the task. 
Both models can be combined using several combination techniques that we review in this section.
	
\begin{table}[!ht]
\centering
\begin{tabular}[t]{l>{\raggedright\arraybackslash}l>{\raggedright\arraybackslash}l>{\raggedright\arraybackslash}p{0.4\linewidth}}
\hline
Shaping method	& Model         & Advice                & References\\
\hline
Reward shaping  & Model-free    & Contextual instructions & \cite{clouse_teaching_1992}\\
                &               & Evaluative feedback   & \cite{isbell_social_2001,thomaz_reinforcement_2006-1,tenorio-gonzalez_dynamic_2010,mathewson_simultaneous_2016}\\

                & Model-based   & Contextual instructions& \cite{najar_socially_2015}\\
                &               & Evaluative feedback   & \cite{knox_combining_2010,knox_augmenting_2011,knox_reinforcement_2012-1}\\

Value shaping   & Model-free    & General instructions          & \cite{utgoff_two_1991,maclin_creating_1996,kuhlmann_guiding_2004,maclin_knowledge-based_2005,maclin_giving_2005,torrey_advice_2008}\\
                &               & Evaluative feedback   & \cite{dorigo_robot_1994,colombetti_behavior_1996,najar_training_2016}\\

                & Model-based   & Contextual instructions& \cite{najar_social-task_2015,najar_training_2016}\\
                &               & Evaluative feedback   & \cite{knox_combining_2010,knox_augmenting_2011,knox_reinforcement_2012-1}\\

Policy shaping  & Model-free    & Contextual instructions& \cite{rosenstein_supervised_2004}\\
                &               & Evaluative feedback   & \cite{ho_teaching_2015,macglashan2017interactive,najar_interactively_20}\\

                & Model-based   & Contextual instructions & \cite{pradyot_integrating_2012,grizou_robot_2013,najar_interactively_20}\\
                &               & Evaluative feedback   & \cite{knox_combining_2010,knox_augmenting_2011,knox_reinforcement_2012-1,lopes_simultaneous_2011,griffith_policy_2013,loftin_learning_2016}\\
                &               & Corrective feedback   & \cite{lopes_simultaneous_2011}\\
Decision biasing&               & Guidance              & \cite{suay_effect_2011,thomaz_reinforcement_2006}\\
                &               & Contextual instructions & \cite{thomaz_robot_2007,tenorio-gonzalez_dynamic_2010,cruz_interactive_2015,nicolescu_natural_2003,rybski_interactive_2007,rosenstein_supervised_2004}\\

\hline
\end{tabular}
\hspace{1cm}
\caption{Shaping methods.} 
\label{tab:shaping} 
\end{table}

\begin{figure}[h!]
\begin{center}
\includegraphics[trim=3.5cm 2.5cm 2cm 2.5cm, clip=true,scale=0.8]{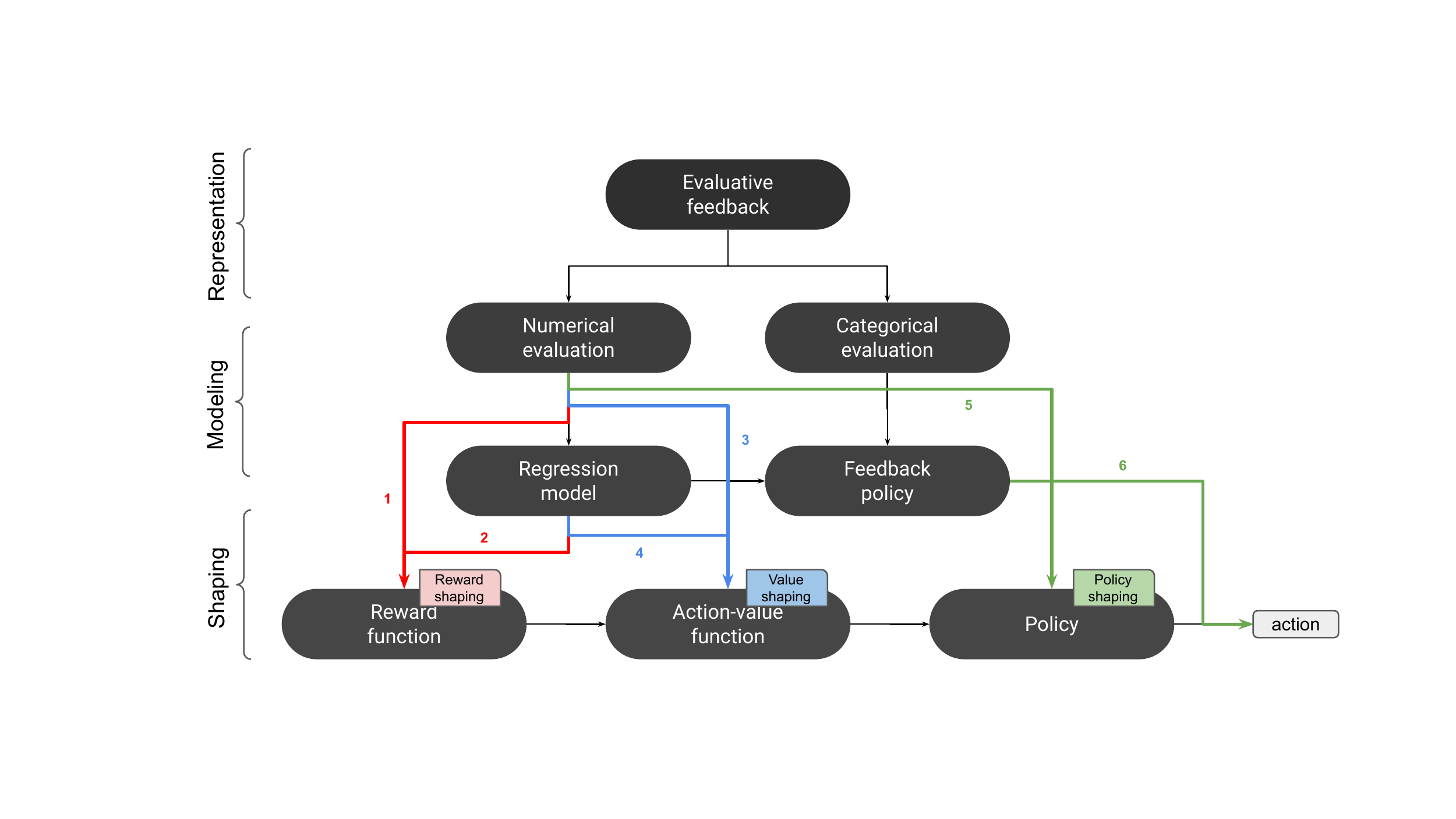}
\caption{Shaping with evaluative feedback. 1: model-free reward shaping. 2: model-based reward shaping. 3: model-free value shaping. 4: model-based values shaping. 5: model-free policy shaping. 6: model-based policy shaping.} 
\label{fig:feedback}
\end{center}
\end{figure}

\textbf{Reward shaping:}
%Reward shaping
%standard reward shaping with feedback
%model-free feedback
Traditionally, reward shaping has been used as a technique for providing a RL agent with intermediate rewards to speed-up the learning process \cite{gullapalli_shaping_1992,mataric_reward_1994,ng_policy_1999,wiewiora_potential-based_2003}. 
One way for providing intermediate rewards is to use evaluative feedback \cite{isbell_social_2001,thomaz_reinforcement_2006-1,tenorio-gonzalez_dynamic_2010,mathewson_simultaneous_2016}. 
In these works, evaluative feedback was considered in the same way as the feedback provided by the agent's environment in RL; so intermediate rewards are homogeneous to MDP rewards.
After converting evaluative feedback into a numerical value, it can be considered as a delayed reward, just like MDP rewards, and used for computing a value function using standard RL algorithms (cf. Fig. \ref{fig:feedback}) \cite{isbell_social_2001,thomaz_reinforcement_2006-1,tenorio-gonzalez_dynamic_2010,mathewson_simultaneous_2016}. 
This means that the effect of the provided feedback extends beyond the last performed action.
When the RL agent has also access to a predefined reward function $R$, a new reward function $R'$ is computed by summing both forms of reward: $R'=R + R^h$, where $R^h$ is the human delivered reward.
This way of shaping with is model-free in that the numerical values provided by the human teacher are directly used for augmenting the reward function.

\begin{figure}[h!]
\begin{center}
\includegraphics[trim=3.5cm 2.5cm 2cm 2.5cm, clip=true,scale=0.8]{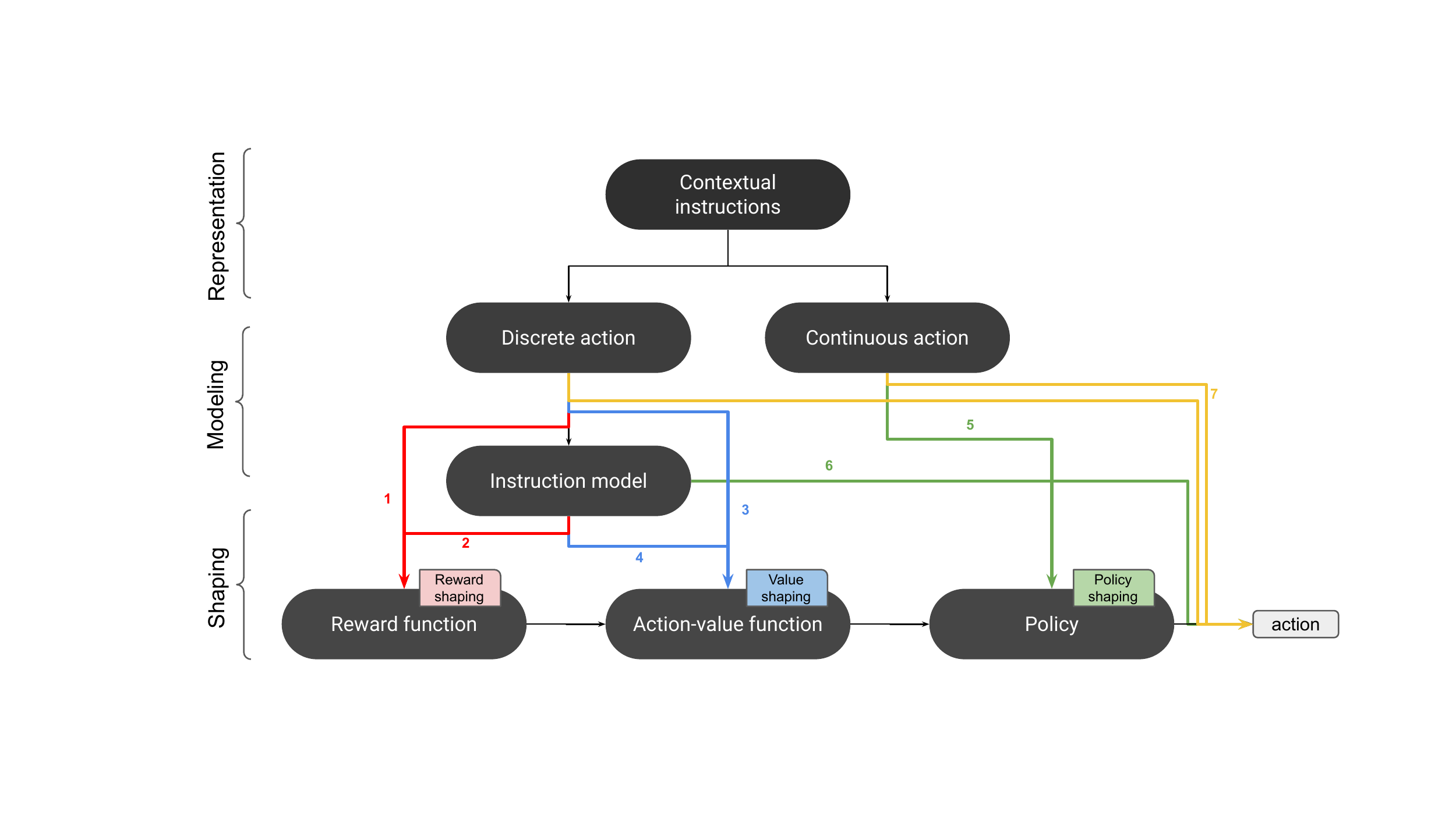}
\caption{Shaping with contextual instructions. 1: model-free reward shaping. 2: model-based reward shaping. 3: model-free value shaping. 4: model-based value shaping. 5: model-free policy shaping. 6: model-based policy shaping. 7: decision biasing.} 
\label{fig:instructions}
\end{center}
\end{figure}

%model-free instructions
Reward shaping can also be performed with instructions (cf. Fig. \ref{fig:instructions}).
For example, in \cite{clouse_teaching_1992}, \textit{contextual instructions} were integrated into a RL algorithm by positively reinforcing the proposed actions in a model-free fashion.

%model-based feedback
%knox
Other works considered building an intermediate model of human rewards to perform model-based reward shaping.
In the TAMER framework \cite{knox_interactively_2009}, evaluative feedback was converted into rewards and used for computing a regression model $\hat{H}$, called the \textit{"Human Reinforcement Function"}.
This model predicted the amount of rewards $\hat{H}(s,a)$ that the human provided for each state-action pair $(s,a)$. 
Knox and Stone \cite{knox_combining_2010,knox_augmenting_2011,knox_reinforcement_2012-1} proposed eight different shaping methods for combining the \textit{human reinforcement function} $\hat{H}$ with a predefined MDP reward function $R$. 
One of them, Reward Shaping, generalizes the reward shaping method by introducing a decaying weight factor $\beta$ that controls the contribution of  $\hat{H}$ over $R$: 

\begin{equation} \label{eq:reward-shaping}
R'(s,a)=R(s,a)+\beta*\hat{H}(s,a).
\end{equation}

%model-based instructions
Model-based reward shaping can also be performed with \textit{contextual instructions}.
In \cite{najar_socially_2015}, a human teacher provided social cues to humanoid robot about the next action to perform.
A model of these cues was built in order to predict task rewards and used simultaneously for reward shaping.

\textbf{Value shaping:}
%Value shaping
%definition: augmenting the value function
%model-free feedback
While investigating reward shaping, some authors pointed out the fundamental difference that exists between immediate and delayed rewards \cite{dorigo_robot_1994,colombetti_behavior_1996,knox_reinforcement_2012}.
Particularly, they considered evaluative feedback as an immediate information about the value of an action, as opposed to standard MDP rewards \cite{ho2017social}. 
For example, in \cite{dorigo_robot_1994}, the authors used a \textit{myopic discounting} scheme by setting the discount factor $\gamma$ to zero. 
This way, evaluative feedback constituted \textit{"immediate reinforcements in response to the actions of the learning agent"}, which comes to consider rewards as equivalent to action values. 
So, value shaping constitutes an alternative to reward shaping by considering evaluative feedback as an action-preference function.
The work of Dorigo and Colombetti \cite{dorigo_robot_1994} was one of the earliest examples of model-free value-shaping.
Another example can be found in \cite{najar_training_2016}, where evaluative feedback was directly used for updating a robot's action values with \textit{myopic discounting}. 

%model-free general advice
Model-free value shaping can also be done with \textit{general advice}.
For example, \textit{if-then} rules can be incorporated into a kernel-based regression model by using
the Knowledge-Based Kernel Regression (KBKR) method \cite{mangasarian_knowledge-based_2004}.
This method was used for integrating \textit{general constraints} into the value function of a SARSA agent using Support Vector Regression for value function approximation \cite{maclin_knowledge-based_2005}.
In this case, advice was provided in the form of constraints on action values (e.g. \textit{if} condition \textit{then} $Q(s,a) \geq1 $), and incorporated into the value function through the KBKR method.
This approach was extended in \cite{maclin_giving_2005}, by proposing a new way of defining constraints on action values.
In the new method, pref-KBKR (preference KBKR), the constraints were expressed in terms of action preferences (e.g. \textit{if} condition \textit{then} prefer action $a$ to action $b$).
This method was also used in \cite{torrey_advice_2008}.
Another possibility is given by the Knowledge-Based Neural Network method (KBANN) which allows incorporating
knowledge expressed in the form of \textit{if-then} rules into a neural network \cite{towell_knowledge-based_1994}.
This method was used in RATLE, an advice-taking system based on Q-learning that used a neural network to approximate its Q-function \cite{maclin_creating_1996}. 
\textit{General instructions} written in the form of \textit{if-then} rules and \textit{while-repeat} loops were incorporated into the Q-function using an extension of KBANN method.
In \cite{kuhlmann_guiding_2004}, a SARSA agent was augmented with an \textit{Advice Unit} that computed additional action values.
\textit{General instructions} sere expressed in a specific formal language in the form of \textit{if-then} rules. 
Each time a rule was activated in a given state, the value of the corresponding action was increased or decreased by a constant in the Advice Unit, depending on whether the rule advised for or against the action.
These values were then used for augmenting the values generated by the agent's value function approximator.

%model-based feedback
Model-based value shaping with evaluative feedback has been investigated by Knox and Stone \cite{knox_reinforcement_2012}, by comparing different discount factors for the \textit{human reinforcement function} $\hat{H}$.
The authors demonstrated that setting the discount factor to zero was better suited, which came to consider $\hat{H}$ as an action-value function more than a reward function\footnote{The authors proposed another mechanism for handling temporal credit assignment, in order to alleviate the effect of highly dynamical tasks \cite{knox_interactively_2009}. In their system, human-generated rewards were distributed backward to previously performed actions within a fixed time window.}. 
The numerical representation of evaluative feedback is used for modifying the Q-function rather than the reward function.
One of the shaping methods that they proposed, Q-Augmentation \cite{knox_combining_2010,knox_augmenting_2011,knox_reinforcement_2012-1} uses the human reinforcement function $\hat{H}$ for augmenting the MDP Q-function using: 

\begin{equation} \label{eq:q-augmentation}
Q'(s,a)=Q(s,a)+\beta*\hat{H}(s,a),
\end{equation}

where $\beta$ is the same decaying weight factor as in Equation \ref{eq:reward-shaping}.

%model-based instructions 
Model-based value-shaping can also be done with \textit{contextual instructions}.
In \cite{najar_social-task_2015,najar_training_2016}, a robot built a model of contextual instructions in order to predict action values, which were used in turn for updating the value function.

\textbf{Policy shaping:}
%Policy shaping
The third shaping strategy is to integrate the advice directly into the agent's policy.
%feedback
%model-free feedback
%It should be noted that in the aforementioned policy shaping methods, evaluative feedback was used only for biasing the MDP policy at decision-time, while reward and value shaping methods modified the task model.
%More recent policy shaping methods take a hybrid approach, where policy shaping is performed by modifying the agent's policy.
Examples of model-free policy shaping with evaluative feedback can be found in \cite{macglashan2017interactive} and \cite{najar_interactively_20}.
In both methods, evaluative feedback was used for updating the actor of an Actor-Critic architecture. 
In \cite{macglashan2017interactive} the update term was scaled by the gradient of the policy:

\begin{equation} 
w \gets w + \alpha \nabla_w \ln \pi_w(a_t|s_t) f_t.
\end{equation}

In \cite{najar_interactively_20}, however, the authors did not consider a multiplying factor for evaluative feedback: 

\begin{equation} 
w \gets w + \alpha f_t.
\end{equation}

%model-free instructions
Model-free policy shaping with \textit{contextual instructions} was considered in \cite{rosenstein_supervised_2004}, in the context of an Actor-Critic architecture, where the error between the instruction and the \textit{actor}'s decision was used as an additional term to the TD error for updating the \textit{actor}'s parameters:

\begin{equation} 
w \leftarrow w + \alpha [k\delta_t (a^E-a^A) + (1-k) (a^S-a^A)] \nabla_w \pi^A(s),
\end{equation}

where $a^E$ is the actor's exploratory action, $a^A$ its deterministic action, $a^S$ the teacher's action, $\pi^A(s)$ the actor's deterministic policy, and $k$ an interpolation parameter.

%model-based feedback
%In policy shaping, evaluative feedback is used for biasing the MDP policy, without interfering with the value function.
Knox and Stone proposed two model-based policy shaping methods for evaluative feedback \cite{knox_combining_2010,knox_augmenting_2011,knox_reinforcement_2012-1}.
Action Biasing uses the same equation as Q-Augmentation (Eq. \ref{eq:q-augmentation}) but only in decision-making, so that the agent's Q-function is not modified: 

\begin{equation} \label{eq:action-biasing}
a^*=argmax_a [Q(s,a)+\beta*\hat{H}(s,a)].
\end{equation}

The second method, Control Sharing, arbitrates between the decisions of both value functions based on a probability criterion.
A parameter $\beta$ is used as a threshold for determining the probability of selecting the decision according to $\hat{H}$:
\begin{equation} \label{eq:control-sharing}
Pr(a=argmax_a[\hat{H}(s,a)])=min(\beta,1).
\end{equation}
Otherwise, the decision is made according to the MDP policy.

Other model-based policy shaping methods do not convert evaluative feedback into a scalar but into a categorical information \cite{lopes_simultaneous_2011,griffith_policy_2013,loftin_learning_2016}.
The distribution of provided feedback is used within a Bayesian framework in order to derive a policy.
The method proposed in \cite{griffith_policy_2013} outperformed Action Biasing, Control Sharing and Reward Shaping.
After inferring the teacher's policy from the feedback distribution, it computed the Bayes optimal combination with the MDP policy by multiplying both probability distributions: $\pi \propto \pi_R \times \pi_F$, where $\pi_R$ is the policy derived from the reward function and $\pi_F$ the policy derived from evaluative feedback.
In \cite{lopes_simultaneous_2011}, both evaluative and corrective feedback were considered under a Bayesian IRL perspective.

%model-based instructions
Model-based policy shaping can also be performed with \textit{contextual instructions}.
For example, in \cite{pradyot_integrating_2012}, the RL agent arbitrates between the action proposed by its Q-learning policy and the one proposed by the instruction model, based on a confidence criterion:

\begin{equation}
\kappa_{\pi}(s) = \max_{a \in A} \pi(s,a) - \max_{b \in A;b \neq a} \pi(s,b).
\end{equation}

The same arbitration criterion was used in \cite{najar_interactively_20}, to decide between the outputs of a an Instruction Model and a Task Model.

%celemin_coach:_2019 ?

%	instructions
% In \cite{pradyot_integrating_2012}, the authors defined two types of contextual instructions: $\pi-$instructions and $\phi-$ instructions.
% $\pi-$instructions set the probability of selecting a particular action to $1$. 
% For example, pointing to an object makes the agent perform a predefined action on it. 
% $\phi-$instructions, on the other hand, reduce the complexity of the current state by projecting its representation into a subspace of features. 
% For example, pointing to an object makes the agent consider only its features and ignore all other aspects.
% This is another way to bias the exploration strategy by focusing on a sub-region of the state-space.

%In \cite{pradyot_integrating_2012}, authors combine both types of instructions in a prior training phase, where a same instruction signal (example pointing to an object) is interpreted using both types of instructions simultaneously and used for computing two separate models, one for each type of instruction. In a later phase, the agent makes its decision in a given time-step by choosing the action of the best interpretation according to a predefined confidence measure over the two instruction models.

\textbf{Decision biasing:}
%decision-biasing
In the previous paragraphs, we said that policy shaping methods can be either model-free, by directly modifying the agent's policy or model-based, by building a model that is used at decision-time to bias the output of the policy.
A different approach consists in using advice to directly bias the output of the policy at decision-time without corrupting the policy nor modelling the advice.
This strategy, that we call decision biasing, is the simplest way of using advice as it only biases the exploration strategy of the agent, without modifying any of its internal variables.
In this case, learning is done indirectly by experiencing the effects of following the advice.

%guidance
%thomaz, chernova, etc.
This strategy has been mainly used in the literature with guidance and contextual instructions.
For example, in \cite{suay_effect_2011,thomaz_reinforcement_2006} guidance reduces the set of actions that the agent can perform at a given time-step.

%instructions
Contextual instructions can also be used for guiding a robot along the learning process \cite{thomaz_robot_2007,tenorio-gonzalez_dynamic_2010,cruz_interactive_2015}. 
%TODO:explain thomaz,tenorio and cruz
For example, in \cite{nicolescu_natural_2003} and \cite{rybski_interactive_2007}, a LfD system was augmented with verbal instructions, in order to make the robot perform some actions during the demonstrations.
In \cite{rosenstein_supervised_2004}, in addition to model-free policy shaping, the provided instruction was also used for decision biasing.
The robot executed a composite real-valued action that was computed as a linear combination of the \textit{actor}'s decision and the supervisor's instruction:

\begin{equation}
a \leftarrow ka^E+(1-k)a^S,
\end{equation}

where $a^E$ is the actor's exploratory action, $a^S$ the supervisor's action and $k$ an interpolation parameter.

%TODO: unclassified
%In \cite{utgoff_two_1991}, the authors presented a State Preference method (SP), where a teacher interactively informed a Temporal Difference (TD) agent about the next preferred state.
%µThis information could be provided by telling the agent what action to perform.
%State preferences were transformed into linear inequalities, which were integrated into the TD algorithm in order to accelerate the learning process.

%\cite{najar_social-task_2015} 

\section{Discussion}
\label{Discussion}

In this section, we first discuss the difference between the various forms of advice introduced in Section \ref{taxonomy}.
We then discuss the approaches presented in Section \ref{interpretation} and Section \ref{shaping}.
Finally, we open some perspectives towards a unified view of interactive learning methods.

\subsection{Comparing different forms of advice}
%Introduction
When designing an advice-taking system, one may ask which type of advice is best suited \cite{suay_practical_2012}.
In this survey, we categorized different forms of advice according to how they are provided to the system.
Even though the same interpretation and shaping methods can be applied to different categories of advice, each form of advice requires a different level of involvement from the human teacher and provides a different level of control over the learning process.
Some of them provide poor information about the policy, so the learning process relies mostly on autonomous exploration.
Others are more informative about the policy, so the learning process mainly depends on the human teacher.

\begin{figure}[h!]
\begin{center}
\includegraphics[width=\textwidth]{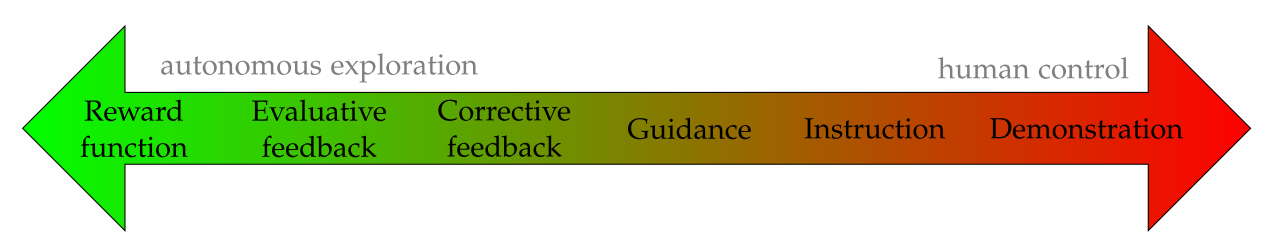}
\caption{Exploration-control spectrum. As we move to the right, teaching signals inform more directly about the optimal policy and provide more control to the human over the learning process.}
\label{fig:exploration-control}
\end{center}
\end{figure}

This aspect has been described in the literature as the guidance-exploration spectrum \cite{breazeal_learning_2008}.
In Section \ref{taxonomy}, we presented guidance as a special type of advice.
So, in order to avoid confusion about the term guidance, we will use the term exploration-control spectrum instead of guidance-exploration (Fig. \ref{fig:exploration-control}).
In the following paragraphs, we compare different forms of advice along this spectrum, by putting them into perspective with respect to other learning schemes such as autonomous learning and learning from demonstration.

\textbf{Autonomous learning:}
At one end of the exploration-control spectrum, autonomous learning methods assume that the robot is able to autonomously evaluate its performance on the task, through a predefined evaluation function, such as a reward function. 
The main advantage of this approach is the autonomy of the learning process. 
The evaluation function being integrated on board, the robot is able to optimize its behaviour without requiring help from a supervisor. 

However, this approach has some limitations when deployed in real-world settings.
First, it is often hard to design, especially in complex environments, an appropriate evaluation function that could anticipate all aspects of a task \cite{kober_reinforcement_2013}.
Second, this approach relies on autonomous exploration, which raises some practical challenges. 
For example, exploring the space of behaviours makes the convergence of the learning process very slow, which limits the feasibility of such approach in complex problems. 
Also, autonomous exploration may lead to dangerous situations. 
So, safety is an important issue that has to be considered when designing autonomous learning systems \cite{garcia_comprehensive_2015}.

\textbf{Evaluative feedback:}
% Training a robot by evaluating its actions can be an alternative solution to the standard RL approach, whenever the implementation of a proper reward function turns out to be challenging \cite{kober_reinforcement_2013}.
% It can also be effective in situations where it is difficult for the teacher to execute demonstrations, and where instructions would require a sophisticated communication channel.
Evaluative feedback constitutes another way to evaluate the agent's performance that has many advantages over predefined reward functions. 
First, like all other types of teaching signals, it can alleviate the limitations of autonomous learning, by allowing faster convergence rates and safer exploration.
Whether it is represented as categorical information \cite{griffith_policy_2013} or as immediate rewards \cite{dorigo_robot_1994}, it provides a more straightforward evaluation of the policy, as it directly informs about the optimality of the performed action \cite{ho_teaching_2015}.
Second, from an engineering point of view, evaluative feedback is generally easier to implement than a reward function.
If designing a proper reward function can be challenging in practice, evaluative feedback generally takes the form of binary values that can be easily implemented \cite{knox_training_2013}.

Nevertheless, the informativeness of evaluative feedback is still limited, as it is only given as a reaction to the agent's actions, without communicating the optimal one. 
So, the agent still needs to explore different actions, with trial-and-error, as in the autonomous learning setting.
The main difference is that exploration is not required any more once the agent tries the optimal action and gets a positive feedback.
So, the trade-off between exploration and exploitation is less tricky to address than in autonomous learning.
The limitation in the informativeness of evaluative feedback can lead to poor performance.
In fact, when it is the only available communicative channel, people tend to use it also as a form of guidance, in order to inform the agent about future actions \cite{thomaz_reinforcement_2006-1}.
This violates the assumption about how evaluative feedback should be used, which affects learning performance.
Performance significantly improves when teachers are provided with an additional communicative channel for guidance \cite{thomaz_reinforcement_2006}.
This reflects the limitations of evaluative feedback and demonstrates that human teachers also need to provide guidance.

\textbf{Corrective feedback:}
One possibility for improving the feedback channel is to allow for corrections and refinements \cite{thomaz_asymmetric_2007}.
Corrective instructions improve the informativeness of evaluative feedback, by allowing the teacher to inform the agent about the optimal action \cite{celemin_coach:_2019}.
Being also reactive to the agent's actions, they still requires exploration.
However, they prevent the agent from waiting until it tries the correct action by its own, so they requires less exploration compared to evaluative feedback.

On the other hand, corrective instructions require more engineering efforts than evaluative feedback, as they are generally more than a binary information.
Since they operate over the action space, they require from the system designer to encode the mapping between contextual instruction signals and their corresponding actions.
%In this aspect, it is homogeneous to contextual instructions as both operate on the same space.

An even more informative form of corrective feedback is provided by corrective demonstrations, which extend beyond correcting one single action to correcting a whole sequence of actions \cite{chernova_interactive_2009}.
Corrective demonstrations operate on the same space as demonstrations, which require more engineering than contextual instructions and also provide more control over the learning process (cf. the paragraph about demonstrations below).

\textbf{Guidance:}
The experiments of Thomaz and Breazeal have shown that human teachers want to provide guidance \cite{thomaz_reinforcement_2006}.
In contrast to feedback, guidance allows the agent to be informed about future aspects of the task, such as the next action to perform (contextual instruction) \cite{cruz_interactive_2015}, an interesting region to explore (demonstration) \cite{subramanian_exploration_2016} or a set of interesting actions to try (guidance) \cite{thomaz_reinforcement_2006}.

Even though guidance requires less exploration compared to feedback by informing about future aspects of the task, the control over the learning process is exerted indirectly through decision biasing (cf. Section \ref{shaping}).
By performing the communicated guidance, the agent does not directly integrate this information as being the optimal behaviour.
Instead, it will be able to learn only through the experienced effects, for example by receiving a reward.
So guidance is only about limiting exploration, without providing full control over the learning process, as it still depends on the evaluation of the performed actions. 

\textbf{Instructions:}
With respect to guidance, instructions inform more directly about the optimal policy, in two main aspects.
First, instructions are a special case of guidance where the teacher communicates only the optimal action. 
Second, the information about the optimal action can be integrated more directly into the learning process via reward shaping, value shaping or policy shaping.

In Section \ref{taxonomy}, we presented two main strategies for providing instructions: providing general instructions in the form of \textit{if-then} rules, or interactively providing contextual instructions as the agent progresses in the task.
The advantage of general instructions is that they do not depend on the dynamics of the task.
Even though in the literature they are generally provided offline prior to the learning process, there is no reason they cannot be integrated at any moment of the task.
For example, in works like \cite{kuhlmann_guiding_2004}, we can imagine that different rules being activated and deactivated at different moments of the task.
Their integration into the learning process will only depend on the validity of their conditions, not on the moment of their activation by the teacher.
This puts less interactive load on the teacher as he/she does not need to stay concentrated in order to provide the correct information at the right moment.

General instructions also present some drawbacks.
First, they can be difficult to formulate.
The teacher needs to gain insight about the task and the environment dynamics, in order to take into account different situations in advance and to formulate relevant rules \cite{kuhlmann_guiding_2004}.
Furthermore, they require to know about the robot's sensors and effectors in order to correctly express the desired behaviours.
So, formulating rules requires expertise about the task, the environment and the robot.
Second, general instructions can be difficult to communicate.
They require either expert programming skills from the teacher or sophisticated natural language understanding capabilities from the agent.

Contextual instructions, on the other hand, communicate a less sophisticated message at a time, which makes them easier to formulate and to provide. 
Compared to general instructions, they only inform about next the action to perform, without expressing the condition, which can be inferred by the agent from the current task state.
However, this makes them more prone to ambiguity.
For instance, writing general instructions by hand allows the teacher to specify the features that are relevant to the application of each rule, \textit{i.e.,} to control generalization.
With contextual instructions, however, generalization has to be inferred by the agent from the context.

Finally, interactively providing instructions makes it easy for the teacher to adapt to changes in the environment's dynamics.
So they provide more control over the learning process with respect to general instructions.
However, this can be challenging in highly dynamical tasks, as the teacher needs a lapse of time to communicate each contextual instruction. 

%A second form of general advice, \textit{general instructions}, inform more directly about the optimal behaviour, compared to general constraints,
%These batch instructions are themselves composed of a sequence of contextual instructions (cf. next paragraph).
%They can be considered as a special form of communicated demonstrations \cite{lin1992self,whitehead_learning_1991}.
%However, unlike \textit{if-then} rules, state information in general instructions can be implicit. 
%For example, they can be implicitly formulated within the expression of the action (e.g. \textit{"Click start, point to search, and then click for files or folders."}) \cite{branavan_reinforcement_2009}.

\textbf{Demonstration:}
% Compared to all forms of advice, demonstrations are on the control end of the spectrum.
% As they require from the teacher to execute the task, they provide more control over the learning process.
Formally, a demonstration is defined as a sequence of state-action pairs representing a trajectory in the task space \cite{argall_survey_2009}.
So, from a strictly formal view, a demonstration is not very different from a general instruction providing a sequence of actions to perform \cite{branavan_reinforcement_2009,vogel_learning_2010}.
The only difference is in the sequence of states that the robot is supposed to experience.
In many LfD settings, such as teleoperation \cite{abbeel_autonomous_2010} and kinesthetic teaching \cite{akgun_trajectories_2012}, the states visited by the robot are controlled by the human. 
So, controlling a robot through these devices can seen as providing a continuous stream of contextual instructions: the commands sent via the joystick or 
the forces exerted on the robot's kinesthetic device.
So the difference between action plans and demonstrations provided under these settings goes beyond their formal definitions as sequences of actions or state-action pairs.

The main difference between demonstrations and general instructions (actually, all forms of advice), is that demonstrations provide control, not only over the learning process, but also over task execution.
When providing demonstrations, the teacher controls the robot joints, so the communicated instruction is systematically executed.
With instructions, however, the robot is in control of it's own actions. 
Even though the instruction can be integrated into the learning process, via any shaping methods, the robot is still free to execute or not the communicated action.

One downside of this control is that demonstrations involve more human load than instructions.
Demonstrations require from the teacher to be active in executing the task, while instructions involve only communication.
This aspect confers some advantages to instructions in that they offer more possibilities in terms of interaction.
Instructions can be provided with different modalities such as speech or gesture, and by using a wider variety of words or signals.
Demonstrations, however, are constrained by the control interface. 
Moreover, demonstrations require continuous focus in providing complete trajectories, while instructions can be sporadic, like with contextual instructions.

Therefore, instructions can be better suited in situations where demonstrations can be difficult to provide.
For example, people with limited autonomy may be unable to demonstrate a task by themselves, or to control a robot's joints.
In these situations, communication is more convenient.
On the other hand, demonstrations are more adapted for highly dynamical tasks and continuous environments, since instructions require some time to be communicated.

%however the parallel does not hold vision, imitation learning, as the mapping between robots-sate/action need to be found

%this is the correspondence problem, which can be considered a problem of interpreting demonstrations.

% In contrast with instructions, they inform about more than one single action, by communicating sequences of state-action pairs.
% However, providing such control requires to overcome the correspondence problem.
% This is generally addressed through teleoperation or kinesthetic teaching.
% These solutions overcome the correspondence problem in two ways.
% First, the state mapping is avoided as the robot experiences its own states.
% Second, the mapping of actions is made by controlling the robot joints, either through an interface, or by exerting forces on the robot's body.
% This can be seen as sending a continuous stream of instructions: the commands sent via the joystick or the forces exerted on the robot's kinesthetic device.
% So, we can consider demonstrations as a sequence of contextual instructions\footnote{This does not hold for the imitation setting, where the mapping between observed and experienced states is not given.}. 

\subsection{Comparing different interpretation methods}
% 	%supervised vs demonstration vs RL
% 		Cost of annotation
% 		cost of providing demonstrations
% 		limitations of reward functions
% 		use of advice for interpreting other advice

In Section \ref{interpretation}, we presented three main approaches for interpreting advice.
The classical approach, supervised interpretation, relies on annotated data for training linguistic parsers. 
Even though this approach can be effective for building systems that are able to take into account natural language advice, they come at the cost of constituting large corpora of language-to-command alignments.

The second approach, grounded interpretation, relaxes this constraint by relying on examples of task executions instead of perfrectly aligned commands. 
This approach is easier to implement by taking advantage of crowd-sourcing platforms like Amazon Mechanical Turk.
Also, the annotation process is facilitated as it can be performed in the reverse order compared to the standard approach.
First, various demonstrations of the task are collected, for example in the form of videos \cite{tellex_understanding_2011,tellex_learning_2014}.
Then, each demonstration is associated to a general instruction.
Even tough this approach is more affordable than standard language-to-command annotation, it still comes at the cost of providing demonstrations, which can be challenging to provide in some contexts, as discussed in the previous section.

The third approach, RL-based interpretation, relaxes these constraints even more by relying only on a predefined performance criterion to guide the interpretation process \cite{branavan_reinforcement_2009,branavan_reading_2010}. 
Some intermediate methods also exists, for example by deriving a reward function from demonstrations and then using a RL algorithm to interpret advice \cite{vogel_learning_2010,tellex_learning_2014}. 
Given that reward functions can also be challenging to design, some methods rely on predefined advice for interpreting other advice \cite{lopes_simultaneous_2011,mathewson_simultaneous_2016,najar_training_2016}, or a combination of both advice and reward functions \cite{mathewson_simultaneous_2016,najar_interactively_20}.

% 	%robustness to erroneous and sparse signals : difference betwene branavan and reward-based updating
Orthogonal to the difference between supervised, grounded, and RL-based interpretation methods, we can distinguish two different strategies for teaching the system how to interpret unlabeled advice.
The first strategy is to teach the system how to interpret advice without using it in parallel for task learning.
% 	%interactive vs non interactive : general advice vs contextual !!
For example, a human can teach an agent how to interpret continuous streams of contextual instructions by using evaluative feedback \cite{mathewson_simultaneous_2016}.
Here, the main task for the agent is to learn how interpret unlabeled instructions, not to use them for learning another task.
Another example is when the agent is first provided with general instructions, either in the form of \textit{if-then} rules or action plans; and then teaching it how to interpret these instructions using either demonstrations \cite{tellex_understanding_2011,macglashan_translating_2014}, evaluative feedback \cite{macglashan_training_2014} or a predefined reward function \cite{branavan_reinforcement_2009,branavan_reading_2010,vogel_learning_2010}.
%In this case, unlabeled advice is provided prior to the learning process.
In this case, even though the agent is allowed to interact with its environment, the main task is still to learn how to interpret advice, not to use it for task learning. 
%The main goal here is only to interpret unlabeled advice, not to use it for learning another task. 
%Another example is to teach 

The second strategy consists in guiding a task-learning process by interactively providing the agent with unlabeled contextual advice. 
In this case, the agent learns how to interpret advice at the same time as it learns to perform the task \cite{grizou_robot_2013,najar_interactively_20}.
For example, in \cite{grizou_robot_2013}, the robot is provided with a set of hypotheses about possible tasks and advice meanings.
The robot then infers the task and advice meanings that are the most coherent with each other and with the history of observed advice signals. 
In \cite{najar_interactively_20}, task rewards are used for grounding the meaning of contextual instructions, which are used in turn for speeding-up the task-learning process.

It is important to understand the difference between these two strategies.
First, when the agent learns how to interpret advice while using it for task learning, we must think about which shaping method to use for integrating the interpreted advice into the task-learning process (cf. Section \ref{shaping}).
Second, when the goal is only to interpret advice, there is no challenge about the optimality nor the sparsity of the unlabeled advice.

With the first strategy, advice cannot be erroneous as it constitutes the reference for the interpretation process.
Even though the methods implementing this strategy do not explicitly assume perfect advice, the robustness of the interpretation methods against inconsistent advice is not systematically investigated.
When advice is also used for task learning, however, we need to take into account whether or not advice is correct with respect to the target task.
For example, in \cite{grizou_robot_2013}, the authors report the performance of their system under erroneous evaluative feedback.
In \cite{najar_interactively_20}, the system is evaluated in simulation against various levels of error for both evaluative feedback and contextual instructions.
Also with the first strategy, advice signals cannot be sparse since they constitute the state-space of the interpretation process. 
For instance, the standard RL methods that have been used for interpreting general instructions \cite{branavan_reinforcement_2009,branavan_reading_2010,vogel_learning_2010} cannot be used for interpreting sparse contextual instructions. 
In these methods, instructions constitute the state-space of an MDP over which the RL algorithm is deployed, so they need to be instantiated on every time-step.
This problem has been addressed in \cite{najar_interactively_20}, where the system was able to interpret sporadic contextual instructions, by using the TD error of the task-learning process.

\subsection{Comparing different shaping methods}

In Section \ref{shaping}, we presented different methods for integrating advice into a RL process: reward shaping, value shaping, policy shaping and decision-biasing.
The standard approach, reward shaping, has been effective in many domains \cite{clouse_teaching_1992,isbell_social_2001,thomaz_reinforcement_2006-1,tenorio-gonzalez_dynamic_2010,mathewson_simultaneous_2016}.
However, this way of providing intermediate rewards has been shown to cause sub-optimal behaviours such as positive circuits \cite{knox_reinforcement_2012,ho_teaching_2015}.
Even though these effects have been mainly studied under the scope of evaluative feedback, they can also be extended to other forms of advice such as instructions, since the positive circuits problem is inherent to the reward shaping scheme regardless of the source of the rewards \cite{mahadevan_automatic_1992,randlov_learning_1998,ng_policy_1999,wiewiora_potential-based_2003}.

Consequently, many authors considered value shaping as an alternative solution to reward shaping \cite{ho2017social,knox_reinforcement_2012-1}. 
However, when comparing different shaping methods for evaluative feedback, Knox and Stone observed that \textit{"the more a technique directly affects action selection, the better it does, and the more it affects the update to the Q function for each transition experience, the worse it does"} \cite{knox_reinforcement_2012-1}.
In fact, this can be explained by the specificity of the Q-function with respect to other preference functions.
Unlike other preference functions (e.g. Advantage function \cite{harmon_advantage_1994}), a Q-function also informs about the proximity to the goal via temporal discounting.
Contextual advice such as evaluative feedback and contextual instructions, however, only inform about local preferences like the last or the next action, without including such information \cite{ho_teaching_2015}.
So, like reward shaping, value shaping with contextual advice may also lead to convergence problems.

Overall, policy shaping methods show better performance compared to other shaping methods \cite{knox_reinforcement_2012-1,griffith_policy_2013, ho_teaching_2015}.
In addition to performance, another advantage of policy shaping is that it is applicable to a wider range of methods that directly derive a policy, without computing a value function or even using rewards.
%Policy shaping can be convenient for policy-based methods which do not compute a value-function.

\subsection{Toward a unified view}
Overall, all forms of advice overcome the limitations of autonomous learning, by providing more control over the learning process.
Since more control comes at the cost of more interaction load, the autonomy of the learning process is important for minimizing the burden on the human teacher.
Consequently, many advice-taking systems combine different learning modalities in order to balance between autonomy and control.
%So, often, different methods are combined within a single framework.
For example, RL can be augmented with evaluative feedback \cite{judah_reinforcement_2010,sridharan_augmented_2011,knox_reinforcement_2012-1}, corrective feedback \cite{celemin_reinforcement_19}, instructions \cite{maclin_creating_1996,kuhlmann_guiding_2004,rosenstein_supervised_2004,pradyot_integrating_2012}, instructions and evaluative feedback \cite{najar_interactively_20}, demonstrations \cite{taylor_integrating_2011,subramanian_exploration_2016}, demonstrations and evaluative feedback \cite{leon_teaching_2011}, or demonstrations, evaluative feedback and instructions \cite{tenorio-gonzalez_dynamic_2010}.
Demonstrations can be augmented with corrective feedback \cite{chernova_interactive_2009,argall_teacher_2011}, instructions \cite{rybski_interactive_2007}, instructions and feedback, both evaluative and corrective \cite{nicolescu_natural_2003}, or with prior RL  \cite{syed_imitation_2007}.
In \cite{waytowich2018cycle}, the authors proposed a framework for combining different learning modalities in a principled way. 
The system could balance autonomy and human control by switching from demonstration to guidance to evaluative feedback, using a set of predefined metrics such as performance.
%Other initiatives have also been proposed in the literature \cite{cederborg_language_2013,waytowich2018cycle}.
%In \cite{cederborg_social_2014}, the authors proposed a mathematical framework for learning from different sources of information. 
%The main idea is to relax the assumptions about the meaning of teaching signals, by taking advantage of the coherence between the different sources of information.

\begin{figure}[h!]
\begin{center}
\includegraphics[trim= 2cm 7.5cm 0cm 4cm, clip=true,width=\textwidth]{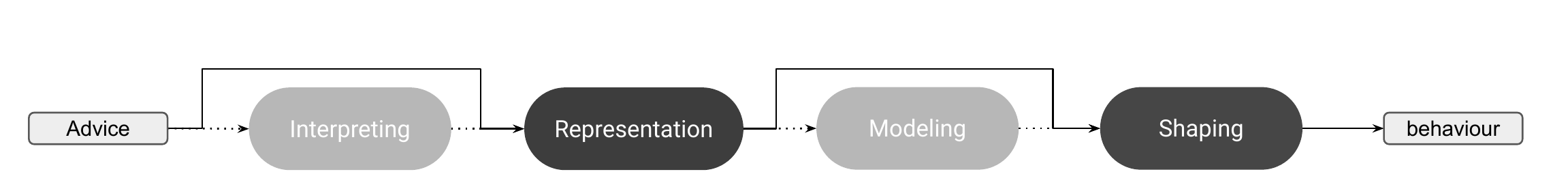}
\caption{Shaping with advice, a unified view. When advice is provided to the learning agent, it has first to be encoded into an appropriate representation. If the mapping between teaching signals and their corresponding internal representation is not predetermined, then advice has to be interpreted by the agent. Then advice can be integrated into the learning process (shaping), either in a model-free or a model-based fashion. Optional steps, interpretation and modeling, are sketched in light grey.
} 
\label{fig:advice}
\end{center}
\end{figure}

Integrating different forms of advice into one single and unified formalism remains an active research question.
So far, different forms of advice have been mainly investigated separately by different communities.
For example, some shaping methods have been designed exclusively for evaluative feedback and were not tested with other forms of advice such as contextual instructions, and the converse is also true.
In this survey, we extracted several aspects that were shared across different forms of advice.
Regardless of the type of advice, we must ask the same computational questions as we go through the same overall process (Fig. \ref{fig:advice}): 
First, we must think about how advice will be represented and whether its meaning will be predetermined or interpreted by the learning agent.
Second, we must decide whether to aggregate advice into a model, or directly use it for influencing the learning process (model-based vs. model-free shaping). 
Finally, we must choose a shaping method for integrating advice (or its model) into the learning process.
From this perspective, all shaping methods that were specifically designed for evaluative feedback could also be used for instructions, and \textit{vice-versa}.
For example, all the methods proposed by Knox and Stone for learning from evaluative feedback \cite{knox_combining_2010,knox_augmenting_2011,knox_reinforcement_2012-1}, can be recycled for learning from instructions. 
Similarly, the confidence criterion used in \cite{pradyot_integrating_2012} for learning from contextual instructions constitutes another Control Sharing mechanism, similar to the one proposed in \cite{knox_combining_2010,knox_augmenting_2011,knox_reinforcement_2012-1} for learning from evaluative feedback.

It is also interesting to think about the relationship between interpretation and shaping. 
For example, we can notice the similarity between interpretation and shaping methods. 
In Section \ref{interpretation}, we mentioned that some interpretation methods relying on the task-learning process can be either reward-based, value-based or policy-based.
This scheme is reminiscent of the different shaping methods: reward shaping, value shaping and policy shaping. 
For instance, the policy shaping method proposed in \cite{griffith_policy_2013} for combining evaluative feedback with a reward function is mathematically equivalent to the Boltzmann Multiplication method used in \cite{najar2017shaping} for interpreting contextual instructions.
So by extension, the other ensemble methods that have been used for interpreting contextual instructions could also be used for shaping. 
We also note that the confidence criterion in \cite{pradyot_integrating_2012} was used for both interpreting instructions and policy shaping.
So, we can think of the relationship between shaping and interpretation as a reciprocal influence scheme, where advice can be interpreted from the task-learning process in a reward-based, value-based or a policy-based way, and in turn can influence the learning process in a reward-based, value-based or policy-based shaping way \cite{najar2017shaping}.
This view contrasts with the standard flow of the advice-taking process, where advice is interpreted before being integrated into the learning process \cite{hayes-roth_advice_1981}.
In fact in many works, interpretation and shaping happen simultaneously, sometimes by using the same mechanisms \cite{pradyot_beyond_2011,najar_imitation_20}.

Under this perspective, we can extend the similarity between all forms of advice to include also other sources of information such as demonstration and reward functions.
At the end, even though these signals can sometimes contradict each other, they globally inform about one same thing, \textit{i.e.,} the task \cite{cederborg_social_2014}.
Until recently, advice and demonstration have been mainly considered as two complementary but distinct approaches, \textit{i.e.,} communication vs. action \cite{dillmann2000learning,argall2008learning,knox_interactively_2009,judah_reinforcement_2010,knox_understanding_2011}.
However, these two approaches share many common aspects.
%For instance, the 5-steps advice-taking process also applies to demonstrations. 
For example, the counterpart of interpreting advice in the LfD literature is the correspondence problem, which is the question of how to map the teacher's states an actions into the agent's own states and actions.
With advice, we also have a correspondence problem that consists in interpreting the raw advice signals.
So, we can consider a more general correspondence problem that consists in interpreting raw teaching signals, independently from their nature.
So far, the correspondence problem has been mainly addressed within the community of learning by imitation. 
Imitation is a special type of social learning in which the agent reproduces what it perceives. 
So, there is an assumption about the fact that what is seen has to be reproduced.
Advice is different from imitation in that the robot has to reproduce what is communicated by the advice and not what is perceived.
For instance, saying "turn left", requires from the robot to perform the action of turning left, not to reproduce the sentence "turn left".
However, evidence from neuroscience gave rise to a new understanding of the emergence of human language as a sophistication of imitation throughout evolution \cite{adornetti_pragmatic_2015}.
In this view, language is grounded in action, just like imitation \cite{corballis_mirror_2010}.
For example, there is evidence that the mirror neurons of monkeys also fire to the sounds of certain actions, such as the tearing of paper or the cracking of nuts \cite{kohler_hearing_2002}, and that spoken phrases about movements of the foot and the hand activate the corresponding mirror-neuron regions of the pre-motor cortex in humans \cite{aziz-zadeh_congruent_2006}.

So, one challenging question it whether we could unify the problem of interpreting any kind of teaching signal under the scope of one general correspondence problem. 
This is a relatively new research question, and few attempts have been made in this direction.
In \cite{cederborg_social_2014}, the authors proposed a mathematical framework for learning from different sources of information.
The main idea is to relax the assumptions about the meaning of teaching signals, by taking advantage of the coherence between the different sources of information.
When comparing demonstrations with instructions, we mentioned that some demonstration settings could be considered as a way of providing continuous streams of contextual instructions, with the subtle difference that demonstrations are systematically executed by the robot. 
Considering this analogy, the growing literature about interpreting instructions \cite{branavan_reading_2010,vogel_learning_2010,grizou_robot_2013,najar_interactively_20} could provide insights for designing new ways of solving the correspondence problem in imitation.

Unifying all types of teaching signals under the same view is a relatively recent research question \cite{cederborg_social_2014,waytowich2018cycle}, and this survey aims at pushing towards this direction by clarifying some of the concepts used in the interactive learning literature and highlighting the similarities that exist between different approaches.
The computational questions covered in this survey extend beyond the boundaries of Artificial Intelligence, as similar research questions regarding the computational implementation of social learning strategies are also addressed by the Cognitive Neuroscience community \cite{biele2011neural,olsson2020neural,najar_imitation_20}.
We hope this survey will contribute in bridging the gap between both communities. 
%Thus we think this survey can be of interest for both communities.

%similarity between interpretation methods and shaping methods: Information from the task such as the reward function, the value function or the policy can be used for interpreting instructions, and in turn the interpretation of advice can be used for shaping by updating one of these elements.

%link between RL-based and demonstration-based interpretation. Grounding instructions with demonstration is based on the same principle as policy-based updating. With demonstrations, the optimal action is given a priori, whereas in RL, the optimal action is learned by triial-and-error.

% all inputs have the same goal: informing about the task, independently from their form. challenge of sparsity and contradiction....

\section{Conclusion}
\label{Conclusion}

In this paper, we provided an overview of the existing methods for integrating human advice into a Reinforcement Learning process. 
We first proposed a taxonomy of the different forms of advice that can be provided to a learning agent.
We then described different methods that can be used for interpreting advice, and for integrating it into the learning process.
Finally, we discussed the different approaches and opened some perspectives towards a unified view of interactive learning methods.

\section*{Acknowledgments}
This work was supported by the Romeo2 project.

\bibliographystyle{apalike}
\bibliography{main.bib}

\end{document}